\documentclass[10pt,journal,compsoc]{IEEEtran}

\usepackage{amsmath,amsfonts}
\usepackage{algorithmic}
\usepackage{algorithm}
\usepackage{array}
\usepackage[caption=false,font=normalsize,labelfont=sf,textfont=sf]{subfig}
\usepackage{textcomp}
\usepackage{stfloats}
\usepackage{url}
\usepackage{verbatim}
\usepackage{graphicx}
\graphicspath{{figures/}{figures/Sup_figs}}
\usepackage{cite}
\usepackage{color}
\usepackage{CJK}
\usepackage{hyperref}
\usepackage{soul}
\usepackage[numbers,sort&compress]{natbib}

\usepackage{bm}
\usepackage{epsfig}
\usepackage{amsmath}
\usepackage{amssymb}
\usepackage{tabularx}
\usepackage{booktabs}
\usepackage{makecell}
\usepackage{mathtools}
\usepackage{multirow}
\usepackage{xcolor}

\def\0{{\bf 0}}
\def\1{{\bf 1}}

\def\etal{{\em et al.}}
\def\eg{{\em e.g.}}
\def\ie{{\em i.e.}}

\def\etal{{\em et al.\/}\,}

\begin{document}
\bstctlcite{IEEEexample:BSTcontrol} 

\title{Unified Multi-Site Multi-Sequence Brain MRI Harmonization Enriched by Biomedical Semantic Style}

\author{Mengqi~Wu, 
Yongheng~Sun,
Qianqian~Wang,
Pew-Thian~Yap,  
Mingxia~Liu,~\IEEEmembership{Senior Member,~IEEE}

\IEEEcompsocitemizethanks{
\IEEEcompsocthanksitem M.~Wu, Y.~Sun, Q.~Wang, P.T.~Yap, and M.~Liu are with the Department of Radiology and Biomedical Research Imaging Center (BRIC), University of North Carolina at Chapel Hill, Chapel Hill, NC 27599, USA. 
M.~Wu and Q.~Wang are also with the Lampe Joint Department of Biomedical Engineering, University of North Carolina at Chapel Hill and North Carolina State University, Chapel Hill, NC 27599, USA. 
\IEEEcompsocthanksitem 
Corresponding author: M.~Liu (Email: mingxia\_liu@med.unc.edu). 
\protect\\
}
}

\IEEEtitleabstractindextext{
\begin{abstract}
Aggregating multi-site brain MRI data can enhance deep learning model training, but also introduces non-biological heterogeneity caused by site-specific variations (\eg,  
differences in scanner vendors, acquisition parameters, and imaging protocols) that can undermine generalizability.
Recent retrospective MRI harmonization seeks to reduce such site effects by standardizing image style (\eg, intensity, contrast, noise patterns) while preserving anatomical content. However, existing methods often rely on limited paired traveling-subject data or fail to effectively disentangle style from anatomy. 
Furthermore, most current approaches address only single-sequence harmonization, restricting their use in real-world settings where multi-sequence MRI is routinely acquired. 
To this end, we introduce MMH, a unified framework for multi-site multi-sequence brain MRI harmonization that leverages biomedical semantic priors for sequence-aware style alignment. 
MMH operates in two stages: (1) 
a diffusion-based global harmonizer that maps MR images to a sequence-specific unified domain using style-agnostic gradient conditioning,  
and (2) a target-specific fine-tuner that adapts globally aligned images to desired target domains. 
A tri-planar attention BiomedCLIP encoder aggregates multi-view embeddings to characterize volumetric style information, allowing explicit disentanglement of image styles from anatomy without requiring paired data. 
Evaluations on 4,163 T1- and T2-weighted MRIs demonstrate MMH’s superior harmonization over state-of-the-art methods in image feature clustering, voxel-level comparison, tissue segmentation, and downstream age and site classification.
\end{abstract}


}

\maketitle

\begin{figure*}
\setlength{\abovecaptionskip}{0pt}
\setlength{\belowcaptionskip}{0pt}
\setlength{\abovedisplayskip}{0pt}
\setlength{\belowdisplayskip}{0pt}
\includegraphics[width=\textwidth]{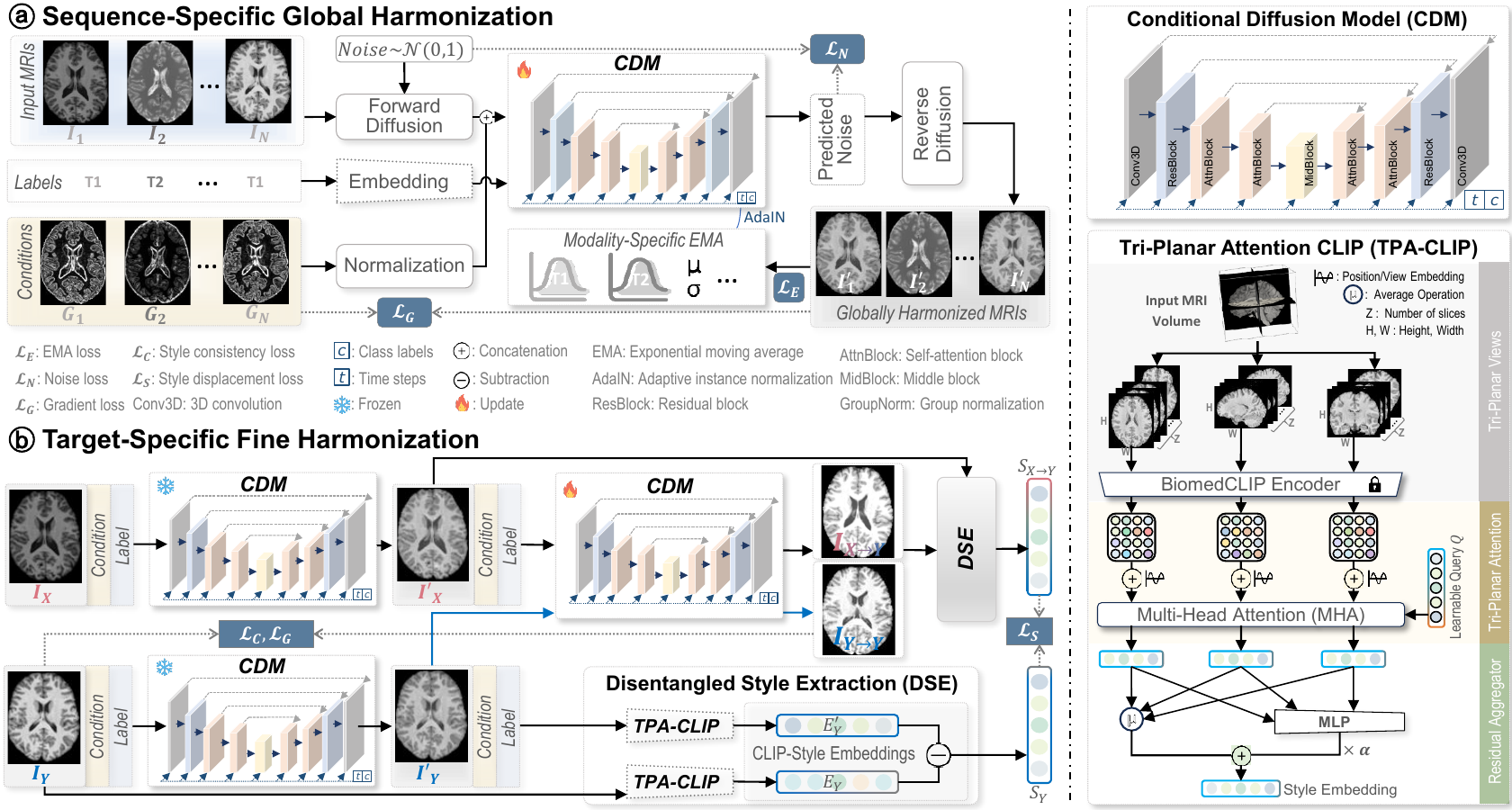}
\caption{Overview of the MMH framework for multi-site multi-sequence brain MRI
harmonization through two-stage learning. (a) Sequence-Specific Global Harmonization: A Conditional Diffusion Model (CDM) is trained to harmonize multi-site, multi-sequence MRIs into unified sequence-specific domains. This stage employs style-agnostic gradient conditioning to anchor anatomy and sequence-specific EMA guidance to eliminate global intensity variations. (b) Target-Specific Fine Harmonization: The pre-trained CDM is fine-tuned to adapt the globally aligned MRIs to a specific target site's style, using a Tri-Planar Attention CLIP (TPA-CLIP) module. We introduce a semantic style displacement strategy, leveraging residuals between raw and globally aligned inputs, to guide robust adaptation without paired training data.}
\label{fig_pipeline}
\end{figure*}

\section{Introduction}
\IEEEPARstart{T}{he} advancement of deep learning (DL) in neuroimaging analysis 
critically depends on large-scale, diverse medical datasets. 
To increase cohort diversity and statistical power, 
studies increasingly rely on aggregating brain MRI data from multiple acquisition sites~\cite{an2022goal,tofts2011multicentre,schnack2010mapping}.
However, this strategy introduces significant {non-biological heterogeneity} arising from differences 
in scanner vendors, field strengths, sequence settings, and acquisition protocols, collectively referred to as \emph{site effects}~\cite{glocker2019machine,wachinger2021detect}. 
Unlike biological variability, site-specific distribution shifts can introduce confounding effects, leading deep learning models to capture spurious correlations rather than robust anatomical or pathological features, and thus compromising their generalizability across acquisition sites~\cite{gadewar2024synthesizing, parida2024quantitative}. 

To mitigate site effects in multi-site studies, retrospective harmonization techniques have emerged as a vital preprocessing step. 
While early statistical-based methods ~\cite{neuroCombat,pomponio2020harmonization} use the empirical Bayes framework to model and correct site effects in the batches of pre-extracted radiomic features, they are limited by their dependency on the feature quality and inability to generalize to unseen data or novel features for diverse downstream tasks~\cite{an2022goal,ImUnity_2023}.
Consequently, image-level methods that harmonize raw image data have become a prevalent strategy for a more generalizable harmonization.

Recent image-level methods have 
shifted toward deep generative models, such as generative adversarial networks (GANs) for image-to-image translation~\cite{StyleGAN,chang_2022_cyclegan,Modanwal_2020_cyclegan}.
However, GAN-based approaches often suffer from training instability, mode collapse, and hallucinations, where the generator alters anatomical structures to satisfy the discriminator. 
Other image-level methods use variational autoencoders (VAEs) or multiple encoder-decoder networks that aim to learn disentangled representations of brain anatomy and image styles~\cite{CALAMITI_2021,dewey2020disentangled,ImUnity_2023}, but they often rely on paired data for training, such as scans from traveling subjects or multiple sequences (\eg, T1- and T2-weighted MRIs) of the same subject to learn complementary information for content and style disentanglement. 
This requirement is often challenging to meet in retrospective studies, posing a practical barrier to their application. 
Diffusion models 
have recently 
been used for MRI harmonization~\cite{durrer2023diffusion,wu2025unpaired}. 
Most of the existing methods cannot jointly harmonize multi-sequence MRIs and require separate training for each sequence. 
This limitation is critical, as multi-sequence data such as T1- and T2-weighted (T1w and T2w) MRIs acquired in the same session often share intrinsic site-specific style characteristics.

To this end, we propose a unified framework for multi-site multi-sequence brain MRI harmonization (\textbf{MMH}) in image space, which leverages biomedical semantic priors to guide MR imaging style alignment. 
As illustrated in Fig.~\ref{fig_pipeline}, our MMH contains two progressive learning steps. 
In MMH, we first train a \emph{global harmonizer} implemented as a site-agnostic conditional diffusion model (CDM), to align multi-sequence MRIs from all sites to sequence-specific unified domains, mitigating global differences (\eg, intensity and contrast). 
This is achieved through the style-agnostic normalized gradient map conditioning and exponential moving average (EMA) constraints that guide global style alignment via adaptive instance normalization (AdaIN)~\cite{huang2017arbitrary}, scalable to multi-sequence input without requiring paired samples.
The second \emph{target-specific harmonization} step fine-tunes the global harmonizer to translate the globally aligned MRIs into a pre-defined target space using a style loss enriched by biomedical image semantics.
To capture spatially complex acquisition styles beyond global intensity characteristics (\eg,~mean, histogram),
a novel tri-planar attention CLIP (TPA-CLIP) module is designed to adapt the pre-trained BiomedCLIP~\cite{zhang2024biomedclip} foundation model for volumetric data by projecting the 3D MRI into three orthogonal views (axial, coronal, sagittal). Unlike simple averaging, TPA-CLIP fuses these multi-view embeddings using a multi-head attention mechanism driven by learnable queries, allowing the model to dynamically attend to informative slice positions while ignoring background noise. Finally, we compute a disentangled ``style vector'' by capturing semantic residual between the original MRI and its globally aligned counterpart, robustly isolating acquisition artifacts from anatomical content.
This dual-stage training scheme leverages the conditional generative power of diffusion models and the semantic-rich embeddings from BiomedCLIP, achieving content-style disentanglement during multi-sequence MRI harmonization without requiring auxiliary encoder-decoder networks or paired training samples. 
The source code is publicly available \href{https://github.com/MW1798/MMH_Harmonization/tree/main}{online}. 

This work presents a substantial extension of our preliminary MICCAI 2025 study~\cite{wu2025unpaired}. 
While our prior method relied on latent diffusion and single-sequence training,
this work advances the framework by:
(1) transitioning to a voxel-space conditional diffusion backbone to eliminate latent compression artifacts and preserve high-frequency details;
(2) establishing a unified multi-sequence framework via sequence-specific EMA constraints to jointly harmonize multi-sequence data (T1w and T2w MRIs);
(3) incorporating normalized gradient-based anatomical anchoring to ensure structural fidelity;
(4) introducing a Tri-Planar Attention (TPA) module to adapt 2D BiomedCLIP for extracting volumetric semantic contexts;
and (5) providing significantly expanded evaluations on a multi-sequence dataset, including region-level analysis, feature clustering, preprocessing robustness, and rigorous ablation studies.

The main contributions of this work that differentiate our prior method~\cite{wu2025unpaired} are summarized as follows:
\begin{itemize}
    \item \textbf{Anatomy-Preserving Gradient Guidance:} We introduce a style-agnostic normalized gradient map conditioning mechanism that decouples anatomical content from imaging style, ensuring robust anatomical preservation during harmonization.

   \item \textbf{3D Biomedical Semantic Guidance:} We introduce a \textbf{Tri-Planar Attention CLIP (TPA-CLIP)} module that adapts the 2D BiomedCLIP foundation model for volumetric data by fusing multi-view embeddings via attention.
   Crucially, we achieve explicit content-style disentanglement by defining the style representation as the \emph{semantic difference} between an image and its globally aligned counterpart, leveraging these high-level semantic priors to guide robust harmonization without paired training subjects.

    \item \textbf{Unified Multi-Sequence Capability:} Unlike single-sequence methods, MMH jointly harmonizes multi-sequence data (\eg, T1w and T2w) within a unified model, demonstrated through extensive validation on 4,163 MRIs across three multi-site datasets.
\end{itemize}

The remainder of this paper is structured as follows. Section~\ref{S_relatedWork}
reviews the most relevant work. Section~\ref{S_method} describes the proposed method. Section~\ref{S_experiment} outlines the experimental setup and
presents the results. Section~\ref{S_discussion} analyzes the impact of several
key components. Section~\ref{S_conclusion} concludes this paper.

\section{Related Work}
\label{S_relatedWork}
\subsection{Non-Learning-Based MRI Harmonization}
Early harmonization efforts primarily addressed site effects through statistical adjustment of tabular data or deterministic intensity transformations. 
Among these, ComBat~\cite{neuroCombat} and its variants~\cite{ComBat_GAM} are the most established, modeling site effects as a combination of multiplicative and additive factors using an empirical Bayes framework. 
While effective for regional analysis, these approaches operate solely on derived scalar features, failing to harmonize raw voxel data required for visualization and downstream tasks~\cite{an2022goal,ImUnity_2023}. 
To address this, signal-based methods perform harmonization directly on voxel intensities.
Approaches like min-max normalization, z-score normalization, and WhiteStripe normalization~\cite{shinohara2014statistical} rescale global voxel intensities to a reference range or intensity distribution.
Histogram matching~\cite{nyul2000new} uses a non-linear approach, warping source intensity histogram to match the cumulative distribution function of a reference target. 
Guan~\etal~\cite{guan2022fast} proposed a frequency-domain harmonization strategy, swapping the low-frequency amplitude spectra (style) of the source with a target while preserving high-frequency details (content). 
These 
methods rely on global intensity statistics or simplified spectral assumptions, failing to account for spatially varying, non-linear variations inherent in multi-site MRI acquisitions. 

\subsection{
Learning-Based MRI Harmonization}
Learning-based harmonization methods generally leverage deep neural networks to directly model complex
and nonlinear site variations, offering greater flexibility in capturing image-level patterns and enabling adaptation to broader downstream tasks. 
Recent advancements cluster into three primary paradigms, including \emph{adversarial learning}, \emph{style-content disentanglement}, and \emph{diffusion-based models} for harmonization, each providing distinct strategies for aligning MRI data across sites while preserving anatomical fidelity.

\subsubsection{Adversarial Learning for  Harmonization}
Generative Adversarial Networks (GANs) pioneered image-level harmonization by modeling site variations as image-to-image translation tasks.
For instance, SiMix~\cite{xu2024simix} utilizes paired traveling subject data and applies cross-site mixing to 
train a pGAN model for direct domain mapping on these synthetically mixed MRIs.
However, the paired traveling subject requirement is impractical in many retrospective studies. 
To address this limitation, CycleGAN-based methods~\cite{CycleGAN2017,chang_2022_cyclegan,Modanwal_2020_cyclegan} introduce cycle-consistency constraints to enable unpaired harmonization.
StyleGAN~\cite{StyleGAN} advanced this by adding a style encoding network to learn a latent style code that can be injected during multi-site mapping.
However, GANs often suffer from unstable training and mode collapse~\cite{croitoru2023diffusion,jung2021conditional}. 
They are also prone to anatomical hallucinations due to the adversarial objective~\cite{cohen2018distribution,kazeminia2020gans}.

\subsubsection{
Style-Content Disentanglement for Harmonization}
Recent studies have shifted toward the disentangled representation learning (DRL) approach, which utilizes encoder-decoder networks to separate anatomical content (\eg,~brain anatomy and tissue boundaries) from image style (\eg,~contrast, intensity, and noise patterns) within a compressed latent space~\cite{dewey2020disentangled,wu2025disentangled,CALAMITI_2021,zuo2023haca3}. 
Early works~\cite{dewey2020disentangled} propose an intra-site supervised approach that leverages paired T1w and T2w MRI data to learn shared anatomical information and reconstruct harmonized MRIs by swapping style latents.
Building upon the paired multi-contrast framework, CALAMITI~\cite{CALAMITI_2021} employs mutual information between anatomy and style, enforcing global disentanglement in latent space. 
HACA3~\cite{zuo2023haca3} extends the sequence-paired paradigm to arbitrary combinations of modalities (\eg, T1w, T2w, and FLAIR) using contrastive learning and an attention-based fusion module.
Conversely, to eliminate paired data reliance, Zuo \etal~\cite{zuo2022disentangling} exploit cross-view self-similarity, using slices from different views to learn consistent contrast with varying content. 
Cackowski~\etal~\cite{ImUnity_2023} propose  ImUnity, which combines a VAE-GAN with an auxiliary site-unlearning network to learn a domain-invariant latent space.
However, most of the existing DRL approaches 
perform 2D slice-level harmonization, neglecting 3D spatial context and causing spatial discontinuity or stripe artifacts~\cite{durrer2023denoising}. 
In addition, their latent space compression inherently incurs high-frequency information loss~\cite{vazquez2024review}, leading to blurred anatomical details.

\subsubsection{Diffusion Models for  Harmonization}
Diffusion models 
have increasingly been used in the medical imaging domain for their stable training objectives and superior generative power over GANs~\cite{croitoru2023diffusion}. 
Durrer~\etal~\cite{durrer2023denoising} train a 2D Denoising Diffusion Probabilistic Model (DDPM) on paired MRIs from traveling subjects for slice-level harmonization.
Our preliminary work~\cite{wu2025unpaired} proposes a 3D latent diffusion model (LDM) for efficient brain MRI harmonization without paired data. We introduce semantic style guidance: unlike traditional methods that rely solely on low-level statistics (\eg,~histograms or Gram matrices), which fail to capture spatially complex acquisition styles, semantic guidance leverages vision-language foundation models to extract high-level site-related styles.
This allows for a more robust disentanglement of the complex, non-linear scanner effects from the underlying biological anatomy.

However, some critical challenges remain among existing approaches.  (1) Methods relying on LDM suffer from the compression artifact, losing high-frequency anatomical details. (2) 
They usually extract semantic features by simply averaging embeddings from 2D axial slices.
This aggregation ignores volumetric correlations, limiting the model's ability to capture view-dependent semantics. (3) Existing frameworks are typically sequence-specific, requiring separate models for harmonizing each MRI sequence. 
To address these limitations, the proposed MMH transitions to a voxel-space conditional diffusion framework to maximize anatomical fidelity and introduces Tri-Planar Attention (TPA-CLIP) to capture robust volumetric semantics.
Furthermore, we unify multi-sequence harmonization within a single model via style-agnostic gradient anchoring and sequence-specific constraints, enabling robust adaptation across diverse MR sequences without requiring paired training data.

\section{Methodology}
\label{S_method}
\subsection{Problem Formulation}
\label{sec:problem_formulation}
In this work, we formulate the multi-site, multi-sequence harmonization as a style transfer problem, where we align the style (\eg,~contrast, intensity, and noise patterns) of MRIs while faithfully maintaining their content (\eg,~brain anatomy). 
Let $\mathcal{X} = \{I_i, m_i, c_i\}_{i=1}^{N}$ represent a heterogeneous dataset of $N$ 3D MRI volumes, where $I_i \in \mathbb{R}^{H \times W \times D}$ denotes the MRI voxel intensity volume; $\small m_i\in\{1,\cdots, M\}$ is a categorical index representing one of $M$ distinct acquisition sequences (\eg, T1w and T2w); and $c_i \in \mathcal{C}$ denotes the site/center label. 
While existing strategies typically aim to either normalize cohorts into a common virtual space~\cite{xu2024simix} or adapt source MRIs to a specific target domain~\cite{ImUnity_2023,zuo2022disentangling,wu2025disentangled}, our MMH framework bridges these objectives through a progressive two-stage training paradigm.

Stage I: Sequence-Specific Global Harmonization. The objective is to learn a mapping function $\Phi_u$ that projects heterogeneous multi-site data into a set of distinct, sequence-specific unified domains $\{\mathcal{U}_1, \cdots, \mathcal{U}_{M}\}$. 
Specifically, the function transforms any input volume $I_i$ into a globally harmonized counterpart $I'_i = \Phi_u(I_i, m_i) \in \mathcal{U}_{m_i}$. 
By conditioning on $m_i$, this formulation ensures that global intensity variations are eliminated within each sequence-specific subspace (\eg, aligning all T1w inputs to a common $\mathcal{U}_{T1}$) while preserving intrinsic contrast differences between sequences. 

Stage II: Target-Specific Fine-Tuning. We aim to learn a refinement mapping $\Phi_Y$ that adapts globally aligned MRIs from any source site, denoted as $X$, to the specific style of the target domain $c_Y\in \mathcal{C}$. Let $I'_X$ represent the globally harmonized version of a source MRI $I_X$. 
The function generates a target-adapted counterpart $I_{X \to Y} = \Phi_Y(I'_X, m, c_Y) \in \mathcal{Y}_{m}$, where $\mathcal{Y}_{m}$ denotes the distribution of sequence $m$ at the target site Y. 
Since our work focuses on the intra-sequence harmonization, this formulation ensures that the translation is sequence-consistent (\eg, mapping $\mathcal{U}_{T1} \to \mathcal{Y}_{T1}$ and $\mathcal{U}_{T2} \to \mathcal{Y}_{T2}$), adopting the target site's acquisition style while preserving the source anatomy.

\subsection{Stage I: Gradient-Anchored Global Harmonization via Sequence-Specific EMA} 
To implement the sequence-specific mapping $\Phi_{u}$ defined in Sec.~\ref{sec:problem_formulation}, we train a single Unified Conditional Diffusion Model (CDM) jointly across all sites and sequences. This framework is governed by three synergistic mechanisms: (1) sequence-specific class embeddings that provide categorical context ($m$); (2) style-agnostic normalized gradient maps, which anchor the generative process to anatomical boundaries independent of imaging contrast, ensuring the anatomical preservation during harmonization; and (3) Exponential Moving Average (EMA) constraints, which serve both as a dynamic harmonization target during optimization and as a global style condition via adaptive instance normalization (AdaIN). 
These components aim to align heterogeneous inputs to stable, sequence-specific unified domains ($\mathcal{U}_m$) while strictly preserving anatomical fidelity. 

\subsubsection{
Image Generation via Gradient-Conditioned Diffusion} 
The backbone of our framework is a 3D conditional diffusion model (CDM)~\cite{ho2020DDPM}. 
The CDM functions as a probabilistic denoiser that learns to progressively reconstruct high-fidelity MRI volumes from random Gaussian noise. This iterative generative process allows the model to effectively capture the complex, high-dimensional distribution of brain anatomy with superior structural fidelity and training stability.
The input is a noisy volume $I^t_i$, generated via a standard forward diffusion process (FDP) that adds scheduled Gaussian noise $\epsilon \sim \mathcal{N}(\mathbf{0}, \mathbf{I})$ to the original MRI $I_i$ over T timesteps, denoted as:
\begin{equation}
\label{eq:fdp}
 I_i^t = \sqrt{\bar\alpha_t}I_i^0 + \sqrt{1-\bar\alpha_t}\epsilon, \quad \epsilon\sim\mathcal{N}(\bm{0},\bm{I}),
\end{equation}
where $\bar\alpha_t$$\coloneqq$$\prod^t_{i=1}\alpha_i$, $\alpha_t$$\coloneqq$$1-\beta_t$, and $\beta_t$ is a predefined variance scheduler~\cite{ho2020DDPM}.
 
To strictly anchor anatomical structures during harmonization, we condition the model on a style-agnostic anatomical prior.
Spatial gradients robustly describe structural geometry by highlighting the high-frequency boundaries between distinct tissue types~\cite{nie2018medical,gonzalez2009digital}.
However, raw gradient magnitudes are inherently coupled with the site-specific contrast.
To decouple geometric content from the imaging style, we compute a percentile-normalized condition $G(I_i)$ by scaling the average spatial gradients by their robust peak magnitude, effectively aligning the dynamic range of edge signals across heterogeneous datasets:
\begin{equation}
\begin{aligned}
    \tilde{G}(I_i) &= Pad(\frac{1}{3}(\nabla_H I_i+\nabla_W I_i+\nabla_D I_i)),\\
    G(I_i) &= \frac{\tilde{G}_i}{Q_{0.99}(|\tilde{G}_i|) + \delta},
\end{aligned}
\end{equation}
where $\nabla$ is the forward-difference operator, $Pad(\cdot)$ restores spatial dimensions through zero padding, $\delta$ is a small constant for numerical stability, and $Q_{0.99}(\cdot)$ is the 99th percentile function. 
By normalizing against this robust peak, we prevent extreme outliers from compressing the informative dynamic range, ensuring a \emph{consistent, site-agnostic anatomical descriptor}. 
The resulting style-agnostic map $G(I_i)$ is concatenated with the noisy image $I_i^t$ as input, while the embedded sequence label $m_i$ is projected into a learnable embedding and added to the timestep embedding $t$, 
providing the necessary categorical context for the conditional generation.

The primary objective of the CDM is implemented as a time-conditioned 3D U-Net to learn a noise prediction network $\epsilon_\theta$. 
This is achieved by minimizing a standard noise-level reconstruction loss~\cite{ho2020DDPM}:
\begin{equation}
\label{eq:noise_loss}
\mathcal{L}_N = \lVert\epsilon - \epsilon_\theta(I_i^t, t, G_i, m_i)\rVert_2^2.
\end{equation}
To enable additional constraints to enforce anatomical integrity and global style alignment, we compute an intermediate one-step estimate of the clean harmonized volume, denoted as $\hat{I}'_i$. 
This can be derived from one reverse diffusion process (RDP), formulated as follows:
\begin{equation}
\label{eq:RDP_onestep}
    \hat{I'_i}\approx I'_i =\frac{1}{\sqrt{\bar\alpha_t}}(I_i^t-\sqrt{1-\bar\alpha_t}\epsilon_\theta(I_i^t, t, G_i, m_i)),
\end{equation}
The estimated image $\hat{I}'_i$ serves as a differentiable proxy for the final globally harmonized output $I'_i$.
We then impose a \emph{gradient consistency loss} $\mathcal{L}_G$ that forces the structural edges of this estimated volume to match the original input's style-agnostic gradient condition $G(I_i)$:
\begin{equation}
\mathcal{L}_G = \lVert G(I_i) - G(\hat{I}'_i) \rVert^2_2.
\end{equation}
This ensures that the harmonization process only alters voxel intensities to match the target style without distorting the underlying anatomical content.

\subsubsection{Global Harmonization via Sequence Style 
Modeling} 
While gradient maps anchor the anatomical content, the harmonization of global image style (\eg, intensity distribution) is achieved via  evolving sequence-specific exponential moving average (EMA) records, denoted as $\mathcal{M}_m = \{ \mathcal{M}_m^{\mathcal{H}}, \mathcal{M}_m^{\mu}, \mathcal{M}_m^{\sigma} \}$, for each sequence $m$, which acts as a dynamic unified domain.

\noindent\textbf{Differentiable Statistic Estimation:}
To maintain these EMA records, we extract differentiable statistics from the intermediate volume $\hat{I}'_i$. 
While the mean ($\mu$) and standard deviation ($\sigma$) are trivially differentiable, capturing the full intensity distribution requires a differentiable soft-histogram. 

Let $x\in\mathbb{R}^F$ be the flattened tensor of $\hat{I'_i}$; we compute a soft-histogram over $K$ bins spanning $[v_{min},v_{max}]$, 
with bin centers defined as $b_k = v_{\text{min}} + \frac{k-1}{K - 1}(v_{\text{max}} - v_{\text{min}})$ for $k=1:K$. 
The contribution of voxel $x_i$ to bin $k$ is computed via a Gaussian kernel: 
\begin{equation}
\small
    w_{ik} = \exp\left(-\frac{1}{2h^2}(x_i - b_k)^2\right),
\end{equation}
where $h$ is the kernel bandwidth controlling smoothness. 
The normalized soft-histogram $\mathcal{H}(x)\in\mathbb{R}^K$ is defined by its $k$-th component as:
\begin{equation}
\label{eq:soft_hist}
\small
    \mathcal{H}_k(x) = \frac{\sum_{i=1}^F w_{ik}}{\sum_{j=1}^{K}(\sum_{i=1}^{F}w_{ij}) + \delta},\quad \text{for } k = 1, \cdots, K,
\end{equation}
where $\delta$ is a stability constant. 
The differentiable soft-histogram of an intermediate volume $\hat{I'_i}$ is denoted as $\mathcal{H}(\hat{I'_i})$. 
We maintain running EMA records of these three statistics ($\mathcal{M}_{m}^{\{\mathcal{H},\mu,\sigma\}}$)
for every sequence type $m$. 
Importantly, these EMA records are not merely passive tracking metrics; they actively guide the generative process. 

\noindent\textbf{Active Style Injection via AdaIN:} 
We employ Adaptive Instance Normalization (AdaIN)~\cite{huang2017arbitrary} to inject the sequence-specific global statistics (mean $\mathcal{M}_m^{\mu}$ and standard deviation $\mathcal{M}_m^{\sigma}$) into the U-Net decoder.
For a feature map $F_l$ at layer $l$, we first normalize it to zero mean and unit variance to remove instance-specific style information~\cite{huang2017arbitrary}. We then modulate it using parameters derived from the global EMA records via a shared multilayer perceptron (MLP):
\begin{equation}
\begin{aligned}
[\boldsymbol{\delta}_m, \boldsymbol{\beta}_m] &= \text{MLP}_l(\mathcal{M}_m^{\mu}, \mathcal{M}_m^{\sigma}), \\
\text{AdaIN}(F_l \mid m) &= (1 + \boldsymbol{\delta}_m) \odot \left( \frac{F_l - \mu(F_l)}{\sigma(F_l)} \right) + \boldsymbol{\beta}_m, 
\end{aligned}
\end{equation}
where $\mu(F_l)$ and $\sigma(F_l)$ are the spatial mean and standard deviation (std.) of the feature map $F_l$. The $\text{MLP}_l$ projects the global style vector into channel-wise scale offsets $\boldsymbol{\delta}_m$ and shift parameters $\boldsymbol{\beta}_m$. Notably, we formulate the scale as a residual term ($1 + \boldsymbol{\delta}_m$) to facilitate identity initialization, ensuring stable gradient flow during the early stages of training.
This explicit conditioning informs the decoder of the current unified intensity distribution for each sequence $m$, \emph{ensuring the generated output aligns globally with the learned sequence-specific domain}.

\noindent\textbf{EMA Record Update and Optimization:}
We update the global EMA records using current batch statistics. However, since the intermediate estimate $\hat{I}'_i$ is derived from the noisy input $I_i^t$ (Eq. \ref{eq:RDP_onestep}), predictions at large timesteps are dominated by Gaussian noise. 
To prevent corrupting the global style prototypes, we implement a timestep-gated update strategy, updating records only when $t$ falls within a reliable threshold $\tau$. 
We compute the soft-histogram for each $\hat{I'_i}$ and update the corresponding sequence-specific record $\mathcal{M}_{m}^{\{\mathcal{H}\}}$: 
\begin{equation}
\label{eq:ema}
\small
    \mathcal{M}_{m}^{\{\mathcal{H}\}} =  \gamma\cdot \mathcal{M}_{m}^{\{\mathcal{H}\}} + (1-\gamma)\mathcal{H}(\hat{I'_i}),~\text{if}~t\leq \tau 
\end{equation}
where $\gamma\in[0,1)$ is the EMA decay factor controlling the update rate. 
We update the $\mathcal{M}_{m}^{\{\mu\}}$ and $\mathcal{M}_{m}^{\{\sigma\}}$ similarly. 
This gating ensures that the global records for histogram ($\mathcal{H}$), mean ($\mu$), and standard deviation ($\sigma$) are only updated when the generative estimate contains meaningful structural information, so that they can serve as the accurate running estimate of the sequence-specific intensity statistics.

After each EMA update, we calculate an EMA consistency loss ($\mathcal{L}_E$) to guide the global harmonization to the unified domain, defined as:
\begin{equation}
\label{eq:loss_ema}
\small
\begin{aligned}
    \mathcal{L}_{E} =& WD(\mathcal{M}_{m}^{\mathcal{H}},\ \mathcal{H}(\hat{I'_i})) \\ &+ \lVert \mathcal{M}_{m}^{\mu} - \mu(\hat{I'_i})\rVert^2_2 + \lVert \mathcal{M}_{m}^{\sigma} - \sigma(\hat{I'_i})\rVert^2_2,
\end{aligned}
\end{equation}
where $WD(\cdot)$ is the differentiable Wasserstein distance, computed as the mean absolute difference between the cumulative distribution functions of two soft-histograms.

\noindent\textbf{Total Objective:} The Stage I loss combines noise reconstruction, anatomical preservation, and global EMA consistency: 
\begin{equation} \mathcal{L}_1 = \mathcal{L}_N + \mathcal{L}_G + \mathcal{L}_E. 
\end{equation}
This formulation aims to train a robust global harmonizer capable of aligning multi-site, multi-sequence MRIs to unified domains without requiring site-specific target references or paired training data. 

\subsection{Stage II: Target-Specific Fine Harmonization}
While Stage I unifies data into generic sequence-specific domains, Stage II aims to adapt these globally aligned representations to the specific acquisition characteristics of a target site $c_Y$. 
We fine-tune the pre-trained CDM using a Tri-Planar Attention (TPA) mechanism that extracts semantic style guidance from a pre-trained BiomedCLIP encoder~\cite{zhang2024biomedclip}.
This semantic guidance augments the low-level EMA style statistics (\eg,~histogram, global mean and standard deviation) with fine-grained acquisition signature (\eg,~scanner-specific contrast and noise patterns) that defines a site's style.
As shown in Fig.~\ref{fig_pipeline}~(b), 
the process translates a globally harmonized source volume $I'_X$ into a target-adapted counterpart $I_{X \to Y}$. 
To achieve this without paired training data, we introduce a novel semantic style displacement strategy by utilizing the globally harmonized volume ($I'$) from Stage I as a canonical content reference for the raw input ($I$).
Since $I$ and $I'$ share identical anatomy but differ in style, the residual between their embeddings isolates a pure ``style vector'', effectively creating a self-supervised style-content pair for every subject, eliminating the need for physical traveling-subject data to disentangle style.


\subsubsection{Target-Oriented Diffusion Fine-Tuning}
We begin by applying global harmonization to raw source $I_X$ and target $I_Y$ inputs using the frozen pre-trained CDM from Stage I ($\epsilon_{\theta}$).
The globally harmonized outputs ($I'_i$ where $i \in \{X, Y\}$) are obtained via an iterative deterministic DDIM~\cite{song2020ddim} sampling strategy ($t=T_r \to 0$):
\begin{equation}
\label{eq:RDP_ddim}
     I^{t-1}_i = \sqrt{\bar\alpha_{t-1}}\hat{I'_i} + \sqrt{1-\bar\alpha_{t-1}}\epsilon_\theta(I_i^t, t, G_i, m_i),
\end{equation}
where $\hat{I'_i}$ is the intermediate estimate used during training (see Eq.~\eqref{eq:RDP_onestep}) and $\epsilon_\theta$ is the trained CDM function.

We then adapt the globally harmonized source MRI $I'_X$ to the target style by fine-tuning a trainable copy of CDM, denoted as $\hat{\epsilon}_\theta$. 
This process is driven by the deterministic DDIM inversion and sampling process~\cite{song2020ddim}. 
Unlike Eq.~\eqref{eq:fdp}, where random noise $\epsilon$ is added to $I_i$, we first invert $I'_X$ into a noisy latent $I^{T_f}_X$ via a deterministic FDP for $T_f$ steps ($t=0:T_f-1$) using CDM-generated noise $\hat{\epsilon}_{\theta}$
\begin{equation}
\label{eq:ddim_fdp}
 I^{t+1}_X = \sqrt{\bar\alpha_{t+1}}\hat{I}^t_X + \sqrt{1-\bar\alpha_{t+1}}\hat{\epsilon}_\theta(I_X^t, t, G_X, m_X),
\end{equation}
We then perform the RDP to denoise over $t=T_r:1$ steps using the fine-tuned model $\hat\epsilon_\theta$ to translate noisy latent $I_X^{T_f}$ back to the final translated MRI  steps: 
\begin{equation}
\label{eq:ddim_rdp}
\small
     I^{t-1}_{X} = \sqrt{\bar\alpha_{t-1}}\hat{I}^t_X + \sqrt{1-\bar\alpha_{t-1}}\hat\epsilon_\theta(I_X^t, t, G_X, m_X),
\end{equation}
where $\hat{I}^t_X$ is the one-step estimate of the clean image at each step $t$ (see Eq.~\eqref{eq:RDP_onestep}).
This generates the target-adapted volume $I_{X \to Y}$, which retains the anatomical structure of $I'_X$ (anchored by gradient maps) but adopts the intensity characteristics of the target domain $Y$ via the semantic style guidance described below.

\subsubsection{Semantic Style Extraction via TPA-CLIP}
To extract semantic style embeddings from an MRI volume, we utilized the pre-trained BiomedCLIP encoder~\cite{zhang2024biomedclip}.
However, standard 2D BiomedCLIP encoders cannot process 3D MRI volumes directly. 
While a prior work~\cite{wu2025unpaired} uses axial slice-averaged embeddings, it disregards the essential 3D spatial context.
To address this, we introduce Tri-Planar Attention CLIP (TPA-CLIP), a module that aggregates semantic features from three orthogonal views (axial, coronal, sagittal) using a multi-head cross-attention mechanism. 

\noindent\textbf{View Extraction and Embedding:}
For a 3D volume, we extract $S_v\in \mathbb{R}^{Z\times H\times W}$ slices along each standard view $v \in \{axial, coronal, sagittal\}$. Each slice is encoded by the pre-trained BiomedCLIP encoder $\Psi$, producing a sequence of slice embeddings $\Psi(S_v) \in \mathbb{R}^{Z \times D}$, where D is the dimension of each embedding vector, and Z is the number of slices.
To preserve spatial location and view identity, we inject sinusoidal positional embeddings $P$ and a learnable view-type embedding $\mathcal{E}_{v}$ to generate a slice-view embedding $\hat{E}_v$:
\begin{equation}
\hat{E}v = \text{Norm}(\Psi(S_v) + P(S_v) + \mathcal{E}_{v}),
\end{equation}where $\text{Norm}$ denotes $L_2$ normalization.


\noindent\textbf{Attention-Based Aggregation:}
To pool the slice-view embedding into a single semantic descriptor $z_v$ of each view, we employ a cross-attention mechanism. For each view $v$, a learnable query vector $Q \in \mathbb{R}^{1 \times D}$ attends to the slice embedding sequence $\hat{E}_v$ via multi-head attention~\cite{vaswani2017attention}: 
\begin{equation}
z_v = \text{LayerNorm}(\text{MHA}(Q, \hat{E}_v, \hat{E}_v)),
\end{equation}
where $Q$ serves as the query and $\hat{E}_v$ acts as both key and value. 
This effectively computes a weighted aggregation of the slice sequence from each view, allowing the model to prioritize semantically informative slices (\eg, those containing distinct anatomical landmarks) while suppressing less relevant ones.

\noindent\textbf{Residual Fusion:}
Finally, the three view descriptors $z_{v\in\{a,c,s\}}$ are fused via a residual MLP aggregator to produce the final semantic embedding $\Psi(V)$:
\begin{equation}
\label{eq:semantic_embedding}
\Psi(V) = \frac{1}{3}\sum_{v\in\{a,c,s\}} z_v + \alpha \cdot \text{MLP}([z_{a}; z_{c}; z_{s}]),
\end{equation}
where $\alpha$ is a learnable scaling factor initialized to 0.1. 

\subsubsection{Disentangled Semantic Style Guidance} 
To guide style translation without requiring paired data or explicit style learning, we propose a \emph{style displacement loss} that explicitly separates "style" from "content" in the semantic space. 
Specifically, we define the style vector $S$ as the difference in semantic embeddings $\Psi(V)$ between an MRI and its globally-aligned counterpart:
\begin{equation}
\label{eq:clip_ebd}
S_{Y} = \Psi(I_Y) - \Psi(I'_Y), \quad S_{X \to Y} = \Psi(I_{X \to Y}) - \Psi(I'_X),
\end{equation}
where $\Psi(\cdot)$ is the semantic image embedding extracted by TPA-CLIP (see Eq.~\eqref{eq:semantic_embedding}). 
Since $I'_Y$ and $I_Y$ share the same brain anatomy of a subject, the residual style vector $S_Y$ captures purely target-specific style, disentangled from content information. 
Similarly, $S_{X \to Y}$ reflects the style vector of the harmonized source image.

We enforce style translation by minimizing the discrepancy between the translated style vector $S_{X\to Y}$ and the target style $S_Y$.
This \emph{style displacement loss} comprises both directional and magnitude constraints:
\begin{equation}
\mathcal{L}_{S} = \Big| \|S_{Y}\|_2 - \|S_{X \to Y}\|_2 \Big| + (1- \frac{S_{Y}\cdot S_{X\to Y}}{\|S_{Y}\|\|S_{X\to Y}\|}), 
\end{equation}
where the 1st magnitude term matches the strength of the style vectors in the CLIP-embedding space, and the 2nd term quantifies the directional discrepancy between two style embeddings. 
To ensure target style consistency, we further design a \emph{style 
consistency loss} by minimizing the distance between style embeddings of each target MRI and its harmonized counterpart: $\small \mathcal{L}_{C} = \left\|S_{Y} - S_{Y\to Y} \right\|_1$. 

The total Stage II objective is a hybrid loss defined as:
\begin{equation}
\label{eq:loss_s2}
    \mathcal{L}_2 = \mathcal{L}_{S} + \mathcal{L}_{C} + \mathcal{L}_G. 
\end{equation}
By exploiting the semantic richness of BiomedCLIP, 
our framework effectively translates source MRIs to the target style without requiring paired data or explicit disentanglement modules. This implicit strategy ensures robust adaptation to the target acquisition characteristics while strictly preserving the anatomical fidelity of the source.

In summary, our progressive dual-stage approach improves the generalizability and flexibility of the framework. 
By decoupling global standardization from target adaptation, the Stage I model functions as a site-agnostic backbone that can process data from unseen sources without retraining. 
Furthermore, adapting to a new target domain requires fine-tuning only the Stage II module, preserving the stable anatomical priors learned in the first stage. 
This simplifies the deployment of MMH to new clinical centers compared to traditional end-to-end retraining.



\begin{figure*}[!tp]
\setlength{\abovecaptionskip}{-2pt} 
\centering
\includegraphics[width=1\textwidth]{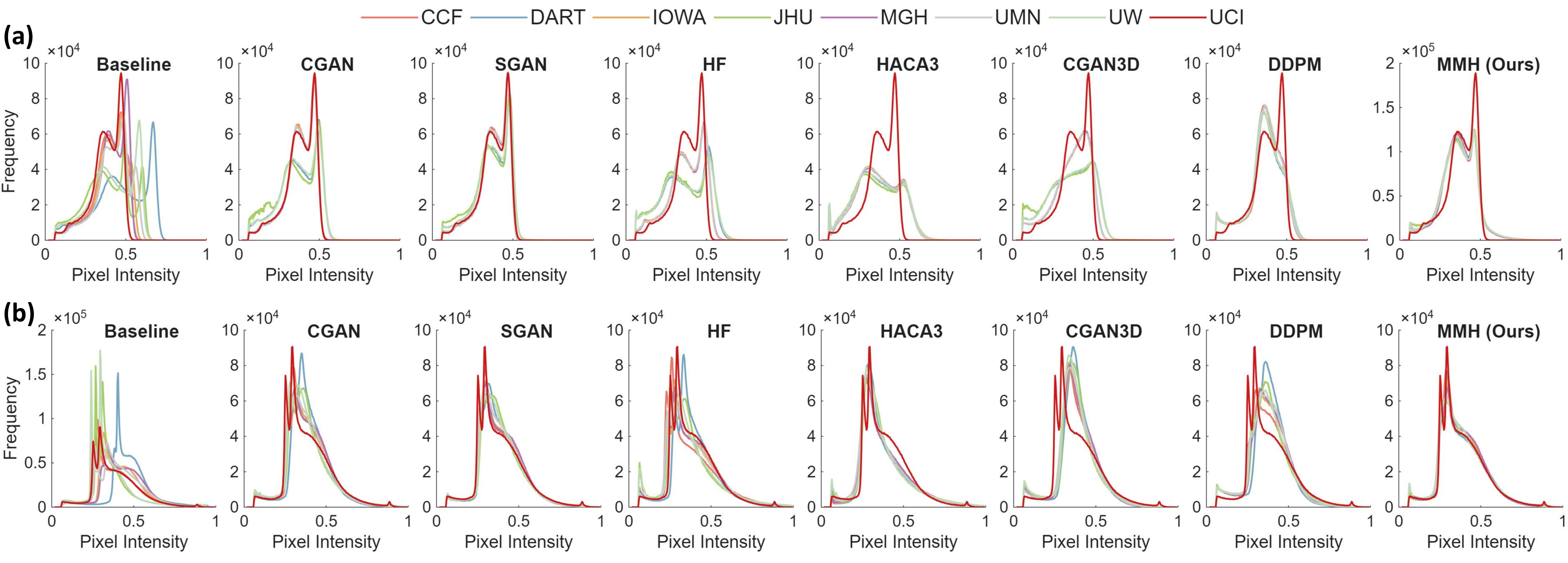}
\caption{Intensity histograms of (a) 16 T1w MRIs and (b) 16 T2w MRIs from the DWI-THP test set across 8 sites. The target domain (UCI) is highlighted in red.}
\label{fig:hist}
\end{figure*}

\subsection{Implementation}
Our MMH is implemented using PyTorch and MONAI~\cite{cardoso2022monai} on an NVIDIA A100 GPU. 
The CDM is implemented as a time-conditioned 3D U-Net with a symmetric architecture. It comprises 2 upsampling/downsampling layers, 2 residual blocks, 4 self-attention blocks, and a middle block, with the channel dimensions \{32, 64, 256, 256\}, respectively. 
For TPA-CLIP, we use a frozen pre-trained BiomedCLIP encoder~\cite{zhang2024biomedclip} with an embedding dimension $D$=$512$. 
The learnable residual aggregator is implemented as a lightweight MLP and initialized with a scaling factor $\alpha$=$0.1$.

Training is performed in two stages using Adam optimizer with an effective batch size of 4 (achieved via gradient accumulation). 
In Stage I, the CDM is trained from scratch with an initial learning rate (LR) of $1\times10^{-4}$.
We employ a linear variance scheduler $\beta$ ranging from $0.0015$ to $0.0195$ over $T$=$1{,}000$ diffusion steps~\cite{ho2020DDPM}.
In Stage II, we fine-tune the CDM and the TPA-CLIP aggregator with a reduced LR of $5$$\times$$10^{-7}$. 
To balance memory constraints with volumetric context during style extraction, we sample $Z$=$24$ equidistant slices per view (ignoring background margins), aggregating $72$ distinct cross-sections per volume.
We set DDIM inversion steps $T_f$=$35$ and sampling steps $T_r$=$25$ via grid search (discussed in Section~\ref{sec:param}).
For the EMA-based style updates, we use a decay factor $\gamma$=$0.8$ and a timestep threshold $\tau$=$100$ to gate updates during noisy intervals. Soft histograms are computed using $K$=$100$ bins over $[v_{\min}, v_{\max}]$ = $[-1, 1]$.
Loss terms are scaled to ensure comparable magnitudes for balanced optimization. See \emph{Supplementary Materials}. 

\begin{figure*}[!tp]
\setlength{\abovecaptionskip}{-2pt}
\setlength{\belowcaptionskip}{0pt}
\setlength{\abovedisplayskip}{0pt}
\setlength{\belowdisplayskip}{0pt}
\centering
\includegraphics[width=1\textwidth]{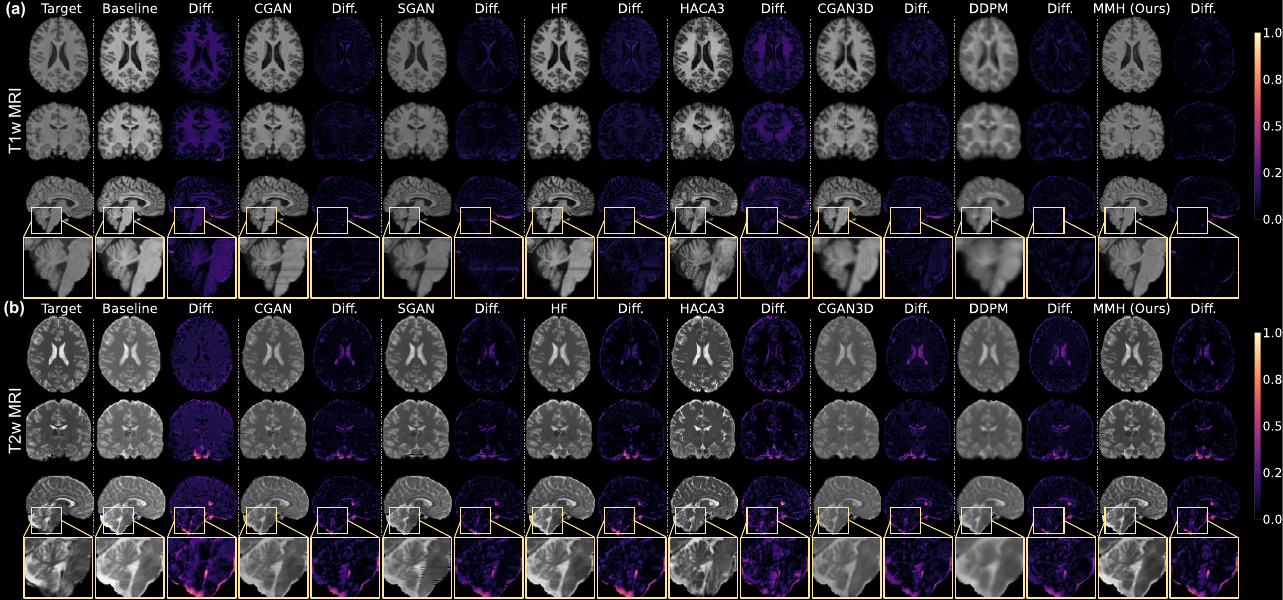}
\caption{
Harmonization results for (a) T1w and (b) T2w MRI from the DWI-THP test set. We display the Target (site UCI), the source MRI (Baseline, site DART), harmonized outputs from six SOTA methods, and our MMH, with absolute difference (Diff.) maps relative to the Target.
}
\label{fig:visual_diff_seg}
\end{figure*}

\begin{figure*}[!t]
\setlength{\abovecaptionskip}{-4pt}
\setlength{\belowcaptionskip}{0pt}
\setlength{\abovedisplayskip}{0pt}
\setlength{\belowdisplayskip}{0pt}
\centering
\includegraphics[width=1\textwidth]{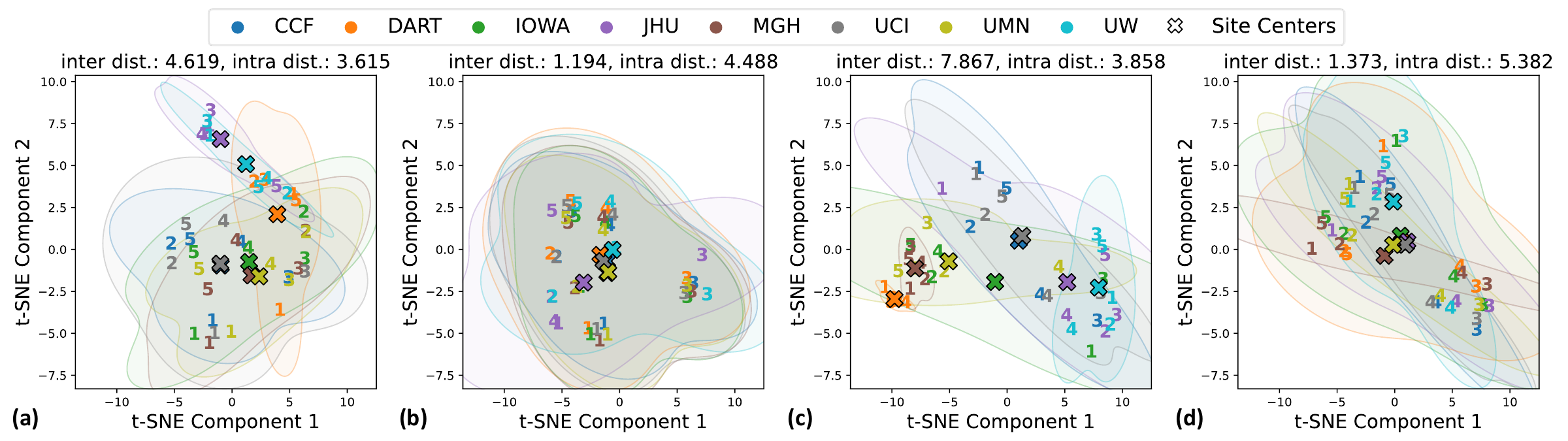}
\caption{
Feature clustering results on (a) 40 unharmonized T1w MRIs; (b) 40 MMH-harmonized T1w MRIs; (c) 40 unharmonized T2w MRIs; and (d) 40 MMH-harmonized T2w MRIs, of 5 subjects across 8 sites in DWI-THP. Numbers denote the subjects, and colors represent the sites.
}
\label{fig:clustering}
\end{figure*}

\begin{figure*}[!tp]
\setlength{\abovecaptionskip}{-4pt}
\setlength{\belowcaptionskip}{0pt}
\setlength{\abovedisplayskip}{0pt}
\setlength{\belowdisplayskip}{0pt}
\centering
\includegraphics[width=1\textwidth]{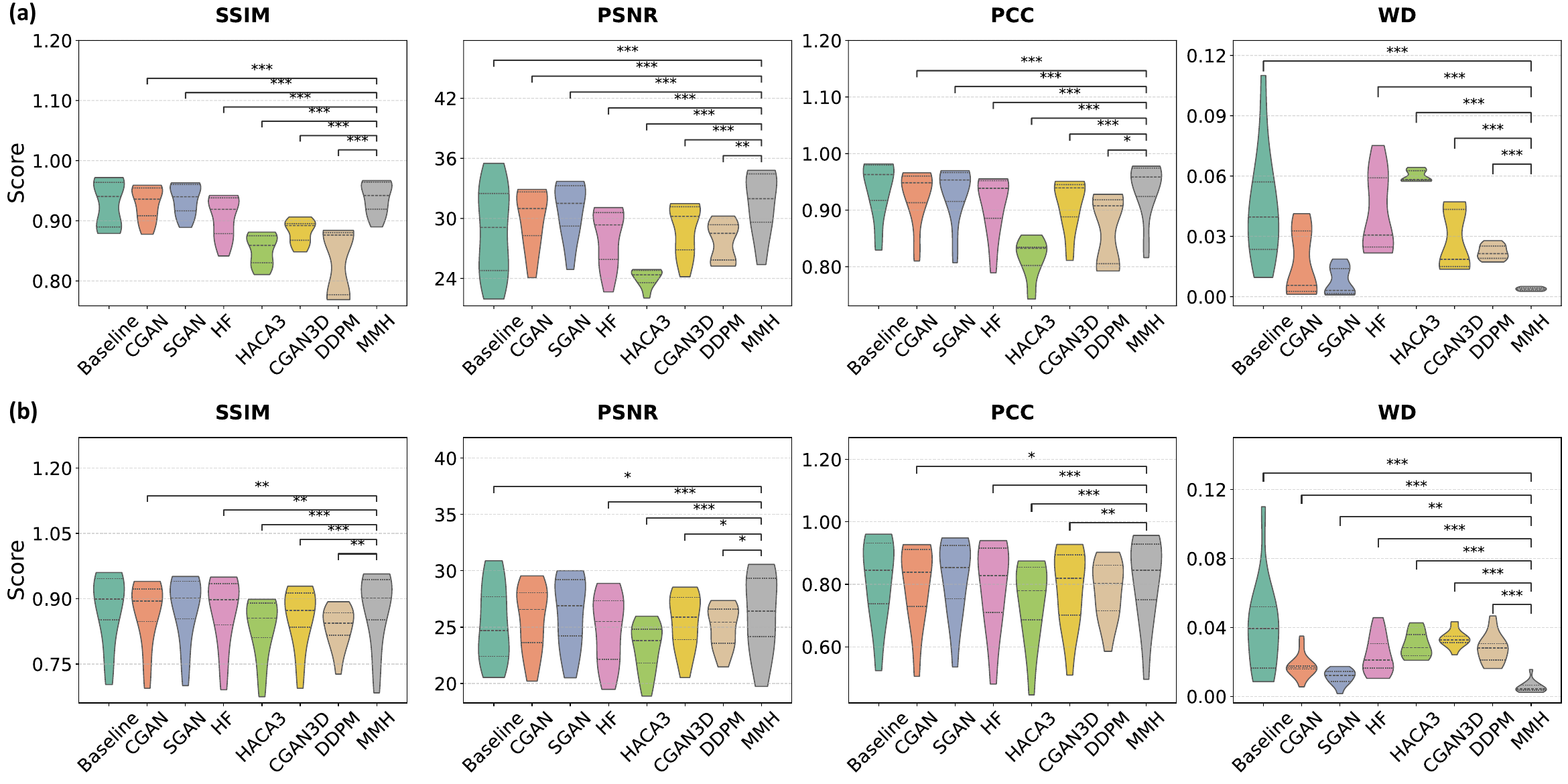}
\caption{Quantitative evaluation of voxel-level harmonization performance on the DWI-THP test set for (a) T1-weighted and (b) T2-weighted MRIs. Metrics (SSIM, PSNR, PCC, WD) compare each harmonized volume against the corresponding traveling subject reference at the target site (UCI). The violin plots illustrate the distribution of scores across the test cohort ($N=16$). Statistical significance between the proposed method (MMH) and competing baselines was assessed using the Wilcoxon signed-rank test ($*$: $p < 0.05$, $**$: $p < 0.01$, $***$: $p < 0.001$).
}
\label{fig:voxel_level}
\end{figure*}

\section{Experiments}
\label{S_experiment}
\subsection{Experimental Setup}
\label{sec:materials}
\noindent\textbf{Datasets}. 
We evaluate our framework using three distinct multi-site MRI datasets, covering large-scale healthy cohorts and paired traveling subjects: 
(1) OpenBHB~\cite{OpenBHB}: A large-scale multi-site dataset containing T1w MRIs from 3,984 healthy subjects acquired across 58 sites. We utilize the official training set comprising 3,227 MRIs; we subdivide it into a training set of 2,259 MRIs (70\%) and a validation split of 968 MRIs (30\%). 
For evaluation, we used the official validation set (757 MRIs) as a standalone test set. 
(2) SRPBS~\cite{SRPBS_TS}: A traveling-subject dataset containing T1w MRIs from 9 participants scanned across 11 sites. We perform a subject-level split to prevent data leakage, using 6 subjects (66 scans) for training, 1 subject (11 scans) for validation, and the remaining 2 subjects (22 scans) for testing. 
(3) DWI-THP~\cite{magnotta2012multicenter}: A multi-sequence dataset containing T1w and T2w MRIs from 5 traveling subjects scanned across 8 sites. We employ a similar subject-level strategy, designating 3 subjects (48 scans) for training with leave-one-subject-out validation, and reserving 2 subjects (32 scans) for testing.

\noindent\textbf{Image Preprocessing and Target Site Selection.} 
We followed the standard T1w MRI preprocessing steps used in existing harmonization literature~\cite{OpenBHB,xu2024simix,StyleGAN,CALAMITI_2021,dewey2019deepharmony,ImUnity_2023}, which include field-of-view (FOV) reorientation, neck cropping, bias field correction, and skull stripping through FreeSurfer~\cite{fischl1999cortical,fischl2012freesurfer}.
All T1w volumes were linearly registered to the MNI-152 1mm template using a 9-degrees-of-freedom (9-DOF) affine transformation. 
Each T2w MRI is linearly co-registered to its corresponding T1w MRI scan. 
Finally, we center-cropped the images to a volume-of-interest (VOI) of $144\times184\times184$ voxels, ensuring complete anatomical coverage while minimizing background margins, and linearly rescaled intensities to the range $[-1, 1]$. 
Given the lack of standard criteria for target site selection in MRI harmonization, 
we formalize the qualitative reference strategy of~\cite{tian2022deep} by selecting the site with the \emph{lowest intra-site style variability}, measured by the mean Wasserstein Distance (WD). 

\noindent\textbf{Competing Methods.}
We compare our MMH with six state-of-the-art (SOTA) image-level MRI harmonization methods:  
CycleGAN (\textbf{CGAN})~\cite{chang_2022_cyclegan},  StyleGAN (\textbf{SGAN})~\cite{StyleGAN}, 
Harmonizing Flow (\textbf{HF})~\cite{beizaee2023HF},
\textbf{HACA3}~\cite{zuo2023haca3},
3D CycleGAN (\textbf{CGAN3D})~\cite{CycleGAN2017},  
and \textbf{DDPM}~\cite{durrer2023diffusion}. 
The 3D methods (\ie,~CGAN3D and DDPM) are trained using the same volumetric dataset as our method, while the 2D methods (\ie,~CGAN, SGAN, and HF) use axial MRI slices derived from the same dataset, ensuring consistent and comparable training hyperparameters for comparisons.
We utilized HACA3's pre-trained weights with the default settings from the official repository. 
We denote the method using raw scans as \textbf{Baseline} in the experiments.

\subsection{Visual and Distributional Qualitative Analysis}
We first assess the MMH's capacity to eliminate global site-specific style disparities while preserving essential brain anatomy. 
Evaluations were conducted on the DWI-THP test set, comprising T1w and T2w MRIs from traveling subjects scanned across 8 sites.
UCI was selected as the target site, with the remaining 7 sites serving as sources. 
Notably, MMH is jointly trained on T1w and T2w data. This parallels HACA3's native multi-sequence capability, whereas all other methods require separate, sequence-specific training runs.
This qualitative assessment focuses on voxel-intensity distribution matching and perceptual image quality.

\subsubsection{Intensity Distribution Alignment Analysis}
As shown in Fig.~\ref{fig:hist}, the Baseline distributions exhibit severe site-wise variations, differing in both global intensity shifts (peaks) and contrast (widths). 
Our MMH effectively aligns the histograms of all source sites to the target (red curve) distribution with high precision. 
In contrast, competing generative models like DDPM and HF homogenize the source sites into a unified distribution that matches the global intensity (peak) but fails to match the variance (histogram width), to the specific target contrast.
Furthermore, despite being trained on a unified multi-sequence manifold, MMH achieves superior histogram overlap across both T1w and T2w modalities. 
This contrasts with the leading single-sequence baselines (\eg, CGAN, SGAN), which, even when trained specifically on independent modality models, fail to achieve consistent alignment across both sequences.

\subsubsection{Perceptual Quality Evaluation}
The visualization results in Fig.~\ref{fig:visual_diff_seg} confirm that MMH-harmonized volumes most closely resemble the target domain appearance for both T1w and T2w sequences. 
The difference maps reveal that MMH minimizes intensity deviation in homogeneous tissue regions while preserving structural edges.
In contrast, the 2D methods (CGAN, SGAN, and HF) introduce noticeable stripe artifacts across different views, indicating limited volumetric consistency.
Conversely, while volumetric baselines (CGAN3D and DDPM) mitigate stripe artifacts, they compromise high-frequency textural fidelity. 
Specifically, CGAN3D exhibited characteristic checkerboard artifacts~\cite{odena2016deconvolution,zhang2018translating}, while DDPM produced over-smoothed outputs due to L2-based objectives inherently favoring average representations~\cite{isola2017image,blau2018perception}.  
These advantages can be traced to the architectural design of MMH. 
Specifically, the improved target-specific alignment stems from the TPA-CLIP module, which models imaging style not merely as low-order statistics (\eg, mean and standard deviation) but as a high-dimensional semantic representation. 
Moreover, the consistent performance across T1w and T2w MRIs 
validates our sequence-specific modulation, which utilizes conditional guidance to maintain distinct, independent style representations for each sequence type. 
Finally, the absence of slice artifacts and preservation of structural edges verify the efficacy of our gradient-based anatomical constraint, which enforces 3D volumetric coherence more rigorously than standard 2D adversarial objectives.

\begin{table*}[!t]
\setlength{\abovecaptionskip}{0pt}
\setlength{\belowcaptionskip}{0pt}
\setlength{\abovedisplayskip}{0pt}
\setlength{\belowdisplayskip}{0pt}
\small
\renewcommand\arraystretch{0.8}
\centering
\caption{Comparison between source site MRIs and corresponding target site (UCI) MRIs with matching traveling subjects on the DWI-THP test set. Asterisk (*) indicates statistically significant differences compared with our method ($p<0.05$, Wilcoxon signed-rank test).}
\label{tab:voxel-metric}
{\small
\begin{tabular}{l|cccc|cccc}
\toprule
\multirow{2}{*}{Method} & \multicolumn{4}{c|}{T1-Weighted MRI} & \multicolumn{4}{c}{T2-Weighted MRI}\\
\cmidrule(lr){2-5} \cmidrule(lr){6-9}
 & SSIM $\uparrow$ & PSNR $\uparrow$ & PCC $\uparrow$ & WD $\downarrow$  & SSIM $\uparrow$ & PSNR $\uparrow$ & PCC $\uparrow$ &WD $\downarrow$  \\
\midrule
Baseline     & $0.930^{0.04}$  & $28.74^{4.37*}$& $\textbf{0.940}^{0.05}$ & $0.044^{0.03*}$& $\textbf{0.879}^{0.08}$  & $25.40^{3.39*}$& $0.818^{0.13}$ &$0.039^{0.03*}$\\
CGAN     & $0.929^{0.03*}$& $30.03^{2.85*}$& $0.926^{0.05*}$& $0.016^{0.02}$& $0.868^{0.07*}$& $25.73^{2.87}$ & $0.803^{0.12*}$&$0.018^{0.01*}$\\
SGAN     & $0.930^{0.03*}$& $27.98^{1.96*}$& $0.930^{0.05*}$& $0.051^{0.01}$  & $0.877^{0.08}$  & $\textbf{26.27}^{3.10}$ & $\textbf{0.821}^{0.12}$ &$0.011^{0.00*}$\\
HF           & $0.906^{0.03*}$& $28.14^{2.79*}$& $0.910^{0.05*}$& $0.041^{0.02*}$& $0.872^{0.08*}$& $24.85^{3.01*}$& $0.795^{0.14*}$&$0.024^{0.01*}$\\
 HACA3& $0.852^{0.02}$  & $24.09^{0.87*}$& $0.815^{0.03*}$& $0.060^{0.00*}$& $0.832^{0.07*}$& $23.16^{2.15*}$& $0.744^{0.13*}$&$0.030^{0.01*}$\\
CGAN3D   & $0.881^{0.02*}$& $28.94^{2.49*}$& $0.911^{0.04*}$& $0.028^{0.01*}$& $0.856^{0.07*}$& $25.41^{2.47*}$& $0.788^{0.12*}$&$0.034^{0.00*}$\\
DDPM         & $0.834^{0.05*}$& $27.66^{1.88*}$& $0.866^{0.06*}$& $0.022^{0.00*}$& $0.834^{0.05*}$& $25.00^{1.84*}$& $0.782^{0.09}$ &$0.028^{0.01*}$\\
MMH (Ours)  & $\textbf{0.938}^{{0.03}}$  & $\textbf{31.52}^{{3.01}}$& $0.938^{{0.05}}$ & $\textbf{0.004}^{{0.00}}$  & $0.877^{{0.08}}$  & $26.20^{{3.48}}$ & $0.815^{{0.13}}$ &$\textbf{0.005}^{{0.00}}$  \\
\bottomrule
\end{tabular}
}
\end{table*}

\subsubsection{Feature Clustering Analysis}
We further investigate the influence of our harmonization on latent image features to verify the effective   disentanglement of site-related variations from anatomical features. 
We utilize a frozen, randomly initialized 3D ResNet18~\cite{cardoso2022monai} with its final classification layer removed to serve as a frozen deep feature extractor. 
We extract image features from DWI-THP MRIs (pre- and post-harmonization by MMH) 
and project them into a shared 2D manifold using the t-SNE~\cite{maaten2008visualizing}. 
In Fig.~\ref{fig:clustering}, colors denote sites, `$\times$' markers indicate site centroids, and numbers (1–5) identify unique traveling subjects.
Unharmonized distributions for T1w and T2w MRIs (Fig.~\ref{fig:clustering} (a) and (c)) reveal distinct site-specific clusters dominated by acquisition bias. 
This is quantified by the large inter-site distance (mean pairwise distance between site centroids), indicating that identical subjects are widely separated solely due to site effects. 
Post-harmonization (Fig.~\ref{fig:clustering} (b) and (d)), two critical changes emerge. 
\emph{First}, site invariance is achieved: the site centroids converge toward a common origin, 
and the site-specific density profiles (colored contours) effectively overlap. 
This visual convergence is supported quantitatively by a substantial reduction in inter-site distances, from $4.62$ to $1.19$ for T1w and from $7.87$ to $1.37$ for T2w, indicating effective removal of global site-related bias. 
\emph{Second}, subject-level structure becomes more pronounced. 
Samples from the same traveling subject (\eg, the clusters labeled `3' and `4') consistently group together across different sites, forming subject-specific sub-clusters. 
At the same time, we observe a modest increase in intra-site distance, measured as the mean distance of samples to their site centroid, reflecting reduced site-driven clustering in favor of biologically meaningful subject-level organization. 
These results demonstrate that MMH effectively removes confounding acquisition styles while preserving subject-specific anatomical identities.

\begin{figure*}[!tp]
\setlength{\abovecaptionskip}{-4pt}
\setlength{\belowcaptionskip}{0pt}
\setlength{\abovedisplayskip}{0pt}
\setlength{\belowdisplayskip}{0pt}
\centering
\includegraphics[width=1\textwidth]{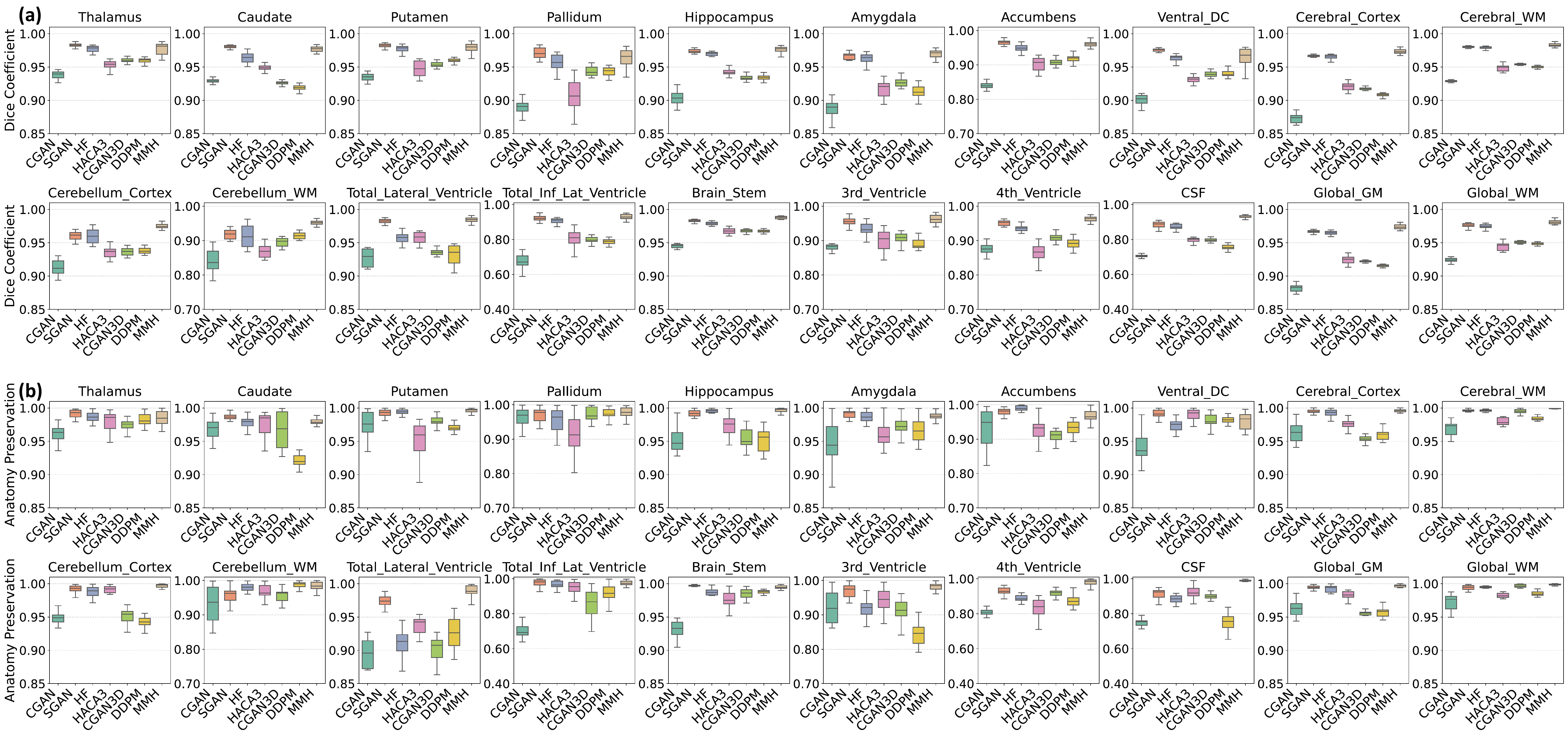}
\caption{Evaluation of anatomy preservation. (a) Dice and (b) Anatomy Preservation scores comparing segmentations of original versus harmonized MRIs across 20 ROIs. Regions are derived from the SynthSeg subcortical atlas, with corresponding left and right structures merged to evaluate bilateral volumetric consistency. These metrics quantify the geometric consistency maintained by each 
method during intensity transformation.}

\label{fig:seg_box}
\end{figure*}

\begin{figure*}[!tp]
\setlength{\abovecaptionskip}{-4pt}
\setlength{\belowcaptionskip}{0pt}
\setlength{\abovedisplayskip}{0pt}
\setlength{\belowdisplayskip}{0pt}
\centering
\includegraphics[width=1\textwidth]{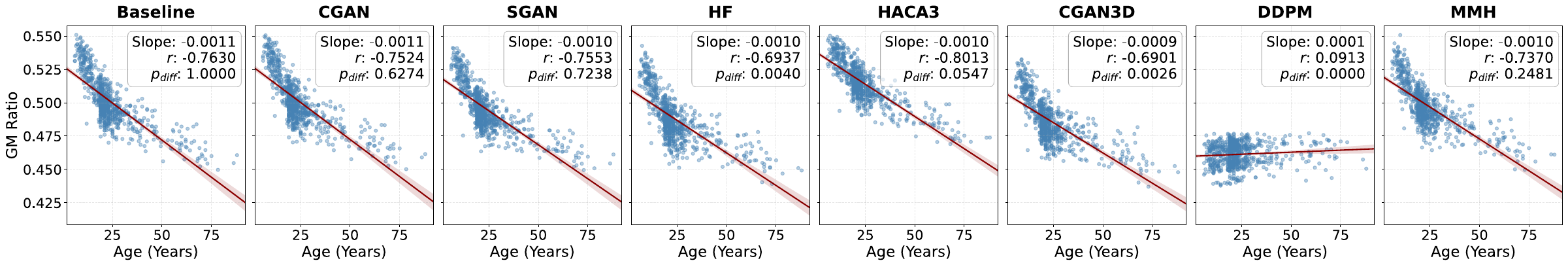}
\caption{Biological consistency evaluation illustrates the relationship between global Gray Matter (GM) ratio and subject age. The red line represents the linear regression fit. Annotations report the regression slope, Pearson correlation coefficient ($r$), and the $p$-value ($p_{diff}$) from Fisher’s $z$-test comparing the correlation of each method against the Baseline. 
} 
\label{fig:age_gm}
\end{figure*}

\subsection{Voxel-Wise Quantitative Analysis}
\label{sec:voxel_wise}
Leveraging the paired traveling-subject MRIs from the DWI-THP test set, we quantitatively compare harmonized MRIs directly against the ground-truth scans at the target site (UCI). 
We utilize the Structural Similarity Index (SSIM), the Peak Signal-to-Noise Ratio (PSNR), and the Pearson Correlation Coefficient (PCC) to measure structural integrity, and
Wasserstein Distance (WD) of intensity histograms to quantify the style alignment. 
Table~\ref{tab:voxel-metric} reports the mean performance, while Fig.~\ref{fig:voxel_level} illustrates the metric distributions.

\subsubsection{Imaging Style Alignment Analysis}
As shown in Table~\ref{tab:voxel-metric}, MMH achieves the lowest WD scores on both sequences (T1w:$0.004$, T2w:$0.005$), significantly outperforming all methods ($p< 0.05$). 
The plots in Fig.~\ref{fig:voxel_level} reveal that the WD distribution of MMH is tightly compressed near zero with low variance. 
In contrast, SGAN and CGAN exhibit broad spreads, indicating high variability in style matching.
Conversely, HACA3 and DDPM show narrow distributions but higher overall WD values, indicating consistent harmonization among sources but a failure to match the specific target style, which corroborates the histogram results in Fig.~\ref{fig:hist}. 
This confirms that MMH yields consistent, target-specific style adaptation across all sequences. 

\noindent\textbf{Structural Fidelity Evaluation}
Table~\ref{tab:voxel-metric} and Fig.~\ref{fig:voxel_level} show that for T1w MRIs, MMH achieves the highest SSIM ($0.938$) and PSNR ($31.52$) alongside superior style alignment (lowest WD). 
Notably, the compact, elevated SSIM distribution (Fig.~\ref{fig:voxel_level}(a)) suggests our gradient-based constraint effectively anchors volumetric generation, preventing distortions common in generative approaches.
For T2-weighted MRIs, MMH maintains anatomical fidelity (second-best SSIM: $0.877$ and PSNR: $26.20$) statistically indistinguishable from the top-performing SGAN and baseline.  
However, SGAN exhibits clearly elevated WD distributions, as a trade-off; thus, MMH matches SGAN's fidelity while reducing style divergence by $50\%$ (WD: $0.005$ vs. $0.011$).
Overall, these results demonstrate that MMH achieves a superior balance, delivering top performance in style harmonization without compromising structural fidelity.

\subsection{Anatomical Segmentation Analysis}
To assess the preservation of brain anatomical structures, we perform automated tissue segmentation on original and harmonized MRIs from the SRPBS test set using SynthSeg~\cite{billot2023robust}. 
We evaluate 20 distinct regions-of-interest (ROIs) based on the SynthSeg look-up table. 
To focus on overall anatomical integrity and reduce regional redundancy, we aggregate the original bilateral labels by merging the left and right components of each ROI (\eg, left and right hippocampuses were combined into a single `Bilateral Hippocampus' ROI).  
This resulted in a total of 20 bilateral structural masks used for the calculation of Anatomy Preservation (AP)  scores~\cite{parida2024quantitative}, 
which measure the relative absolute difference in tissue volumes, 
and Dice Similarity Coefficient (DSC), which quantifies the spatial overlap between segmentation maps. 

\subsubsection{Region-Specific Segmentation Evaluation}
Figure~\ref{fig:seg_box} illustrates DSC and AP distributions across 20 ROIs; MMH consistently exhibits the highest median scores with tight inter-quartile ranges. 
\textbf{(1) Brain aging-related ROIs:} In regions structurally sensitive to age-related atrophy, such as the Hippocampus and Cerebral Cortex, MMH achieves state-of-the-art segmentation (DSC: $97.4\%$ and $97.1\%$, respectively).
Unlike generative baselines (\eg, CGAN, HACA3, and DDPM) that introduce smoothing artifacts in the medial temporal lobe, MMH maintains the high-frequency edge details necessary to distinguish true tissue boundaries, preventing over-harmonization. 
\textbf{(2) Fine-Grained and Complex Geometries:} 
MMH excels in structurally complex regions, most notably in the Cerebellum and subcortical ROIs. 
In the Cerebellum, MMH surpasses the leading slice-based competitor (SGAN) by 1.2\% in the Cerebellar Cortex ($97.3\%$ vs. $96.1\%$) and by 3.0\% in the Cerebellar White Matter ($94.8\%$ vs. $91.8\%$). 
For small anatomical structures like the Amygdala and Brain Stem, MMH maintains a superior mean DSC above $96\%$. 
This improvement 
validates that our 3D gradient constraint successfully resolves spatial coherence often distorted by 2D methods. 

\subsubsection{Global 
Segmentation Evaluation}
At the global level, MMH achieves exceptional volumetric stability, particularly in Global White Matter (GWM), reaching an AP of $99.8\% \pm 0.001$. 
Regarding Subcortical regions (Subc.), while MMH ranks second in mean DSC ($97.2\%$), slightly behind SGAN ($97.4\%$), this difference is statistically insignificant ($p > 0.05$). 
MMH achieves this with a significantly lower standard deviation ($\pm 0.008$ vs. $\pm 0.015$), demonstrating superior robustness against the inconsistencies of adversarial approaches. 
Overall, our method excels in preserving both ROI-level and global-level anatomy while effectively harmonizing MRI styles.
These results can be attributed to our TPA-CLIP's semantic style disentanglement and the normalized gradient constraint, which explicitly anchors the generated volume to source anatomy.

\begin{table}[!t]
\setlength{\abovecaptionskip}{0pt}
\setlength{\belowcaptionskip}{0pt}
\setlength{\abovedisplayskip}{0pt}
\setlength{\belowdisplayskip}{0pt}
\small
\renewcommand\arraystretch{0.8}
\centering
\caption{Anatomy preservation results (\%) on the SRPBS 
dataset. GGM, GWM, and Subc. denote Global Gray Matter, Global White Matter, and Subcortical regions, respectively. Asterisk (*) indicates statistically significant differences compared with our method ($p<0.05$, Wilcoxon signed-rank test).}
\label{tab:ap_dice}
\small
\setlength{\tabcolsep}{0.8pt} 
\begin{tabular}{l|ccc|ccc}
\toprule
\multirow{2}{*}{Method} & \multicolumn{3}{c|}{AP $\uparrow$} & \multicolumn{3}{c}{DSC $\uparrow$} \\
\cmidrule(lr){2-4} \cmidrule(lr){5-7}
 & GGM & GWM & Subc. & GGM & GWM & Subc. \\
\midrule
CGAN    & $96.3^{.012*}$& $97.1^{.012*}$& $95.2^{.015*}$& $88.3^{.005*}$& $92.4^{.003*}$& $90.8^{.031*}$\\
SGAN    & $99.4^{.003}$& $99.4^{.005*}$& $\textbf{98.8}^{.009}$ & $96.6^{.004}$& $97.6^{.003*}$& $\textbf{97.4}^{.008}$ \\
HF      & $99.2^{.005}$& $99.5^{.001}$ & $98.3^{.011}$ & $96.5^{.004*}$& $97.5^{.003*}$& $96.5^{.010*}$\\
HACA3   & $98.2^{.006*}$& $98.2^{.004*}$& $96.1^{.026*}$& $92.0^{.019*}$& $94.1^{.016*}$& $93.2^{.022*}$\\
CGAN3D  & $95.5^{.005*}$& $99.6^{.003*}$& $96.5^{.021*}$& $92.2^{.002*}$& $95.1^{.002*}$& $93.9^{.018*}$\\
DDPM    & $95.7^{.007*}$& $98.5^{.003*}$& $96.1^{.023*}$& $91.5^{.002*}$& $94.8^{.002*}$& $93.9^{.019*}$\\
\textbf{MMH} & $\textbf{99.6}^{.003}$ & $\textbf{99.8}^{.001}$ & $98.5^{.010}$ & $\textbf{97.2}^{.008}$ & $\textbf{98.0}^{.006}$ & $97.2^{.009}$ \\
\bottomrule
\end{tabular}
\end{table}

\subsection{Downstream Evaluation}
To validate the practical utility of our harmonization framework, we perform three downstream tasks using the OpenBHB test set ($N=757$ from 58 sites). 
We treat the raw MRIs as the \textbf{Baseline} and harmonize all volumes to the target domain (Site 17) using each method.
This evaluation is designed to test two complementary objectives: biological consistency (preserving known age-related biological features) and harmonization efficacy (eliminating scanner-specific non-biological features).

\subsubsection{Biological Consistency Evaluation}
We first assess whether the harmonization preserves the well-established neurobiological trend of linear gray matter (GM) atrophy in healthy aging~\cite{pomponio2020harmonization,ge2002age}. 
We employ SynthSeg~\cite{billot2023robust} to generate automated tissue segmentation for each MRI volume and calculate the Gray Matter Ratio by normalizing the Global GM volume by the Total Intracranial Volume (TIV). 
To rigorously quantify biological trend preservation, we perform a linear least-squares regression between subject age and GM ratio and visualize the slope consistency. 
We compare the Pearson Correlation Coefficient of each method ($r_{method}$) against the Baseline ($r_{baseline}$) using Fisher's z-test, which calculates a two-tailed $p$-value ($p_{diff}$) to determine if the harmonization induced a statistically significant deviation in the original biological relationship. 
As shown in Fig.~\ref{fig:age_gm}, MMH, along with CGAN and SGAN, successfully maintains the negative linear correlation of normal brain atrophy and aging trend.
While generative variations cause minor numerical drops in $r$, Fisher's $z$-test confirms these are not statistically significant ($p_{diff} > 0.05$).
In contrast, DDPM, HF, and CGAN3D exhibit significantly degraded correlations while HACA3 yields significantly increased correlations ($p_{diff} < 0.05$), indicating that they alter GM anatomy in original volumes.
This failure is pronounced in DDPM, where the absence of structural anchoring, unlike our gradient constraint, causes structural hallucination or blurred cortical boundaries.

\begin{table}[!t]
\setlength{\abovecaptionskip}{0pt}
\setlength{\belowcaptionskip}{0pt}
\setlength{\abovedisplayskip}{0pt}
\setlength{\belowdisplayskip}{0pt}
\small
\renewcommand\arraystretch{0.8}
\centering
\caption{Site classification and age prediction results on original (Baseline) and harmonized MRIs on OpenBHB.}
\label{tab:site_age_cl}
\setlength{\tabcolsep}{0.5pt}
\small
\begin{tabular}{l|cccc|cc}
\toprule
\multirow{2}{*}{Method} & \multicolumn{4}{c|}{Site Classification (\%)} & \multicolumn{2}{c}{Age Prediction} \\
\cmidrule{2-5} \cmidrule{6-7}
& BACC $\downarrow$ & F1 $\downarrow$ & PRE $\downarrow$ & Recall $\downarrow$ & MAE $\downarrow$ & MSE $\downarrow$ \\
\midrule
Baseline        & $34.3^{2.4}*$& $66.3^{2.3*}$& $75.7^{1.8*}$& $73.2^{1.9*}$& $5.30^{0.26}$ & $47.4^{1.4}$ \\
CGAN            & $42.5^{1.7}$ & $69.5^{2.8*}$& $77.0^{3.0*}$& $73.9^{2.0*}$& $6.63^{0.26*}$& $79.0^{10.5*}$\\
SGAN            & $25.8^{2.2*}$& $59.3^{1.2*}$& $66.2^{1.5*}$& $65.1^{1.5*}$& $7.31^{0.49*}$& $85.7^{12.9*}$\\
HF              & $34.2^{1.1*}$& $66.5^{2.0*}$& $73.6^{2.1*}$& $72.3^{2.1*}$& $5.84^{0.22}$ & $57.0^{3.9*}$\\
HACA3           & $25.7^{2.8*}$& $59.6^{1.4*}$& $68.0^{1.2*}$& $66.2^{1.4*}$& $6.93^{0.63*}$& $74.5^{12.0*}$\\
CGAN3D          & $32.4^{2.9*}$& $65.6^{1.9*}$& $75.1^{1.9*}$& $72.3^{1.7*}$& $5.90^{0.36}$ & $\textbf{43.3}^{7.8}$ \\
DDPM            & $16.6^{1.6*}$& $46.0^{1.3*}$& $54.5^{0.7}$ & $53.5^{2.0*}$& $5.33^{0.26}$ & $47.5^{7.6}$ \\
\textbf{MMH}    & $\textbf{15.3}^{0.5}$ & $\textbf{24.0}^{2.4}$ & $\textbf{38.0}^{3.0}$ & $\textbf{35.3}^{2.4}$ & $\textbf{5.22}^{0.54}$ & $46.7^{8.4}$ \\
\bottomrule
\end{tabular}
\end{table}

\subsubsection{Brain Age Prediction for Clinical Utility Evaluation}
While Fig.~\ref{fig:age_gm} confirms preservation of GM volume and age correlation, we further verify the preservation of the complex, non-linear anatomical features required for clinical inference. 
We utilize a frozen, randomly initialized 3D ResNet18~\cite{cardoso2022monai} to extract high-dimensional features from harmonized MRIs, capturing complex non-linear anatomical descriptors without introducing the inductive biases associated with supervised pre-training. 
We train a ridge regressor on 70\% of the extracted features to predict biological age on the remaining 30\%.  
As reported in Table~\ref{tab:site_age_cl}, MMH achieves the lowest Mean Absolute Error (MAE=5.22), marginally outperforming the unharmonized Baseline (5.30), while competing methods like CGAN (6.63) and SGAN (7.31) significantly degrade performance. 
Notably, DDPM recovers performance in this task (MAE=5.33), as deep feature extractors are robust to the local structural shift. 
However, MMH still demonstrates superior preservation of both the high-level semantic features needed for prediction and the low-level volumetric fidelity required for segmentation.

\subsubsection{Site Classification for Style Removal Evaluation}
Using the same procedure, we train a multi-class logistic regressor to classify the 58 sites. We report Balanced Accuracy (BACC), F1 score (F1), Precision (PRE), and Recall. 
Lower classifier performance indicates successful removal of site-related style features that allow the classifier to distinguish between sites/settings.
Results in Table~\ref{tab:site_age_cl} show that MMH-harmonized MRIs lead to the lowest site classification performance, indicating effective harmonization. 
Importantly, unlike DDPM, which has a low BACC ($16.6$) but higher age prediction error (MAE=$5.33$), MMH effectively removes site features while preserving biological information. 

\begin{table}[!t]
\setlength{\abovecaptionskip}{0pt}
\setlength{\belowcaptionskip}{0pt}
\setlength{\abovedisplayskip}{0pt}
\setlength{\belowdisplayskip}{0pt}
\small
\renewcommand\arraystretch{0.7}
\centering
\caption{Comparison between source site MRIs and corresponding target site (UCI) MRIs with matching traveling subjects on the DWI-THP test set. Asterisk (*) indicates statistically significant differences compared with our method ($p<0.05$, Wilcoxon signed-rank test).}
\label{tab:abl}
\setlength{\tabcolsep}{3pt} 
{\small
\begin{tabular}{l|cccc}
\toprule
Method & SSIM $\uparrow$ & PSNR $\uparrow$ & PCC $\uparrow$ & WD $\downarrow$  \\
\midrule

MMH w/o E & $0.857^{0.09}$ & $27.86^{4.42}$ & $0.869^{0.11}$ & $0.018^{0.01}$ \\
MMH w/o G & $0.829^{0.09}$ & $27.56^{4.32}$ & $0.870^{0.12}$ & $0.016^{0.01}$ \\
MMH w/o T & $0.798^{0.11}$ & $26.50^{3.21}$ & $0.861^{0.11}$ & $0.026^{0.02}$ \\
MMH w/o B & $0.866^{0.08}$ & $27.49^{4.02}$ & $0.867^{0.11}$ & $0.013^{0.01}$ \\
MMH w/o A & $0.823^{0.07}$ & $27.06^{3.82}$ & $0.857^{0.12}$ & $0.012^{0.01}$ \\
\textbf{MMH (Ours)} & $\textbf{0.897}^{0.07}$ & $\textbf{28.38}^{4.62}$ & $\textbf{0.877}^{0.12}$ & $\textbf{0.005}^{0.00}$ \\
\bottomrule
\end{tabular}
}
\end{table}

\begin{figure*}[!tp]
\setlength{\abovecaptionskip}{-2pt}
\setlength{\belowcaptionskip}{0pt}
\setlength{\abovedisplayskip}{0pt}
\setlength{\belowdisplayskip}{0pt}
\centering
\includegraphics[width=1\textwidth]{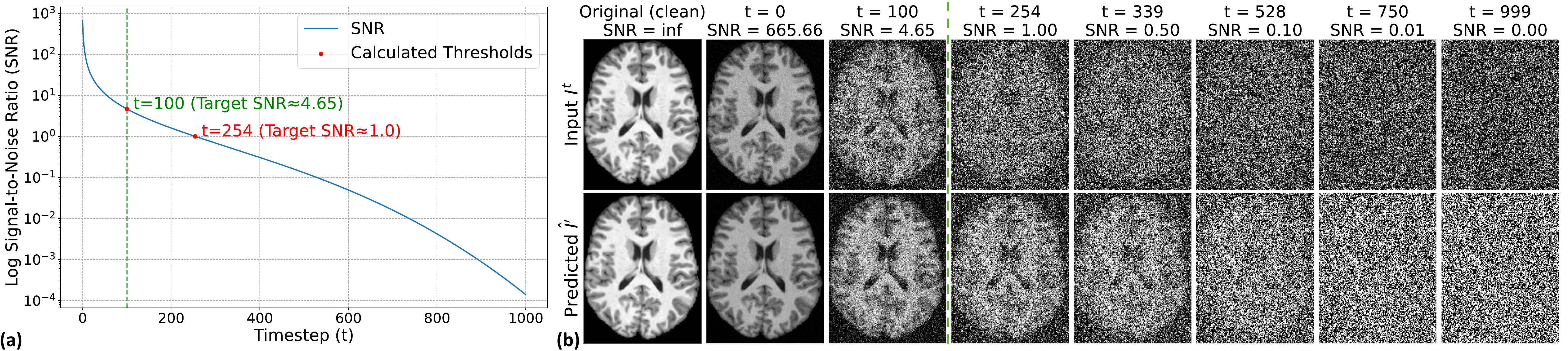}
\caption{Selection of Gating Threshold $\tau$. (a) SNR decay showing the signal-noise crossover at $t \approx 254$ (red dot). (b) One-step reconstructions ($\hat{I}'$) reveal structural loss starting at $t=100$ (green dotted line). We set $\tau=100$ to restrict style updates to this High-Fidelity Regime ($SNR \approx 4.65$), preventing optimization on noise-dominated features.
}
\label{fig:t_th}
\end{figure*}

\section{Discussion}
\label{S_discussion}
\subsection{Ablation Study}
To validate the contributions of individual architectural components, we conduct an ablation study on the DWI-THP test set following the quantitative protocol described in Section~\ref{sec:voxel_wise}. 
Table~\ref{tab:abl} reports aggregated performance (averaged across all subjects and modalities) against its five ablated variants: 
(1) \textbf{MMHw/oE} that removes the EMA-based sequence-specific style modulation to test the necessity of conditional style separation, 
(2) \textbf{MMHw/oG} that excludes the gradient-based anatomical constraint to validate its role in preserving structural edges,  
(3) \textbf{MMHw/oT} that omits the Stage II target adaptation phase (i.e., relies solely on Stage I global harmonization) to quantify the benefit of the target-specific fine-tuning, 
(4) \textbf{MMHw/oB} that replaces the domain-specific \textit{BiomedCLIP} encoder with the standard \textit{CLIP}~\cite{CLIP} pre-trained on natural images, evaluating the impact of medical domain knowledge, and 
(5) \textbf{MMHw/oA} that replaces the Tri-Planar Attention CLIP (TPA-CLIP) module with simple global averaging of axial-slice embeddings, testing the efficacy of our 3D-aware embedding aggregation.

Table~\ref{tab:abl} confirms the contribution of each architectural component.  
\emph{First}, omitting Stage II fine-tuning (MMHw/oT) yields the most severe degradation (WD $\to$ 0.026), confirming target-specific adaptation is the primary driver of domain alignment. 
\emph{Second}, the performance drop in {MMHw/oG} indicates that standard diffusion objectives struggle with high-frequency boundaries; our normalized gradient constraint effectively ``anchors" the generation to the true anatomical prior, improving the anatomical preservation.
\emph{Third}, 
removing the EMA modulation in MMHw/oE causes a $3.6\times$ spike in style divergence, proving that sequence-specific EMA records are required to prevent style conflict between T1w and T2w modalities.
\emph{Fourth}, the failure of {MMHw/oA} highlights the necessity of 3D-aware semantic aggregation; simple axial averaging destroys the volumetric context captured by TPA-CLIP.
\emph{Finally}, the degradation observed in {MMHw/oB} is instructive. 
Replacing BiomedCLIP~\cite{zhang2024biomedclip} with the standard CLIP encoder~\cite{CLIP} (pre-trained on natural images) significantly reduces style alignment. 
This suggests that generic natural-image encoders, whether VGG, ResNet, or standard CLIP~\cite{huang2017arbitrary,gatys2016style,wang2023stylediffusion}, typically prioritize low-level texture statistics (\eg, Gram matrices) over high-level medical domain-specific understanding. 
Whereas, BiomedCLIP possesses ``radiological context'' learned from 15 million medical image-text pairs~\cite{zhang2024biomedclip}, allowing MMH to disentangle site-specific acquisition noise from biological signals more effectively.

\begin{figure}[!t]
\centering
\includegraphics[width=0.99\columnwidth]{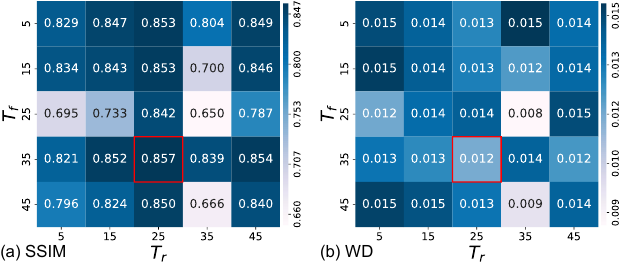}
\caption{Results of MMH with different DDIM sampling parameters.}
\label{fig:para}
\end{figure}

\subsection{Sensitivity to Diffusion Timestep Parameters}
\label{sec:param}
We further analyze two critical temporal mechanisms that govern the framework's stability and expressivity: the loss-gating threshold ($\tau$) used during global pre-training, and the sampling trajectory ($T_f, T_r$) used during harmonization. 

\subsubsection{Training Stability Analysis}
Stage I relies on the accurate estimation of sequence-specific style statistics (soft-histogram $\mathcal{H}$, $\mu$, $\sigma$) for EMA updates (Eqs.~\ref{eq:ema} and \ref{eq:loss_ema}). 
However, these are derived from the one-step estimate $\hat{I}'_i$ (Eq.~\ref{eq:RDP_onestep}). As the diffusion timestep $t$ increases, $\hat{I}'_i$ relies increasingly on the noise prediction $\epsilon_\theta$, leading to inevitable structural degradation. Updating EMA records or calculating consistency losses ($\mathcal{L}_E$) in high-noise regimes risks the ``leakage'' of Gaussian noise properties into the global style prototypes. 
To determine the optimal gating threshold $\tau$, we analyze the Signal-to-Noise Ratio (SNR) decay of our diffusion scheduler in  Fig.~\ref{fig:t_th}(a).
While the theoretical crossover ($SNR=1$) occurs at $t \approx 254$, relying on this upper bound is insufficient for capturing the fine-grained intensity distributions required for the soft-histogram calculation. 
As evidenced by the visual reconstructions of $\hat{I}'_i$ in Fig.~\ref{fig:t_th}(b), the structural boundaries defining distinct tissue intensities degrade perceptibly as early as $t=100$, even while the SNR remains high ($\approx 4.65$).
Consequently, we set a conservative threshold of $\tau=100$. 
This restricts EMA updates and style consistency objectives to the High-Fidelity Regime (SNR $\ge 4.65$), ensuring global style prototypes are constructed solely from structurally reliable estimates and preventing optimization on hallucinatory content.

\subsubsection{Influence of DDIM Sampling Configuration}
DDIM inversion ($T_f$) and sampling ($T_r$) steps control the trade-off between removing source style and anatomical preservation.
Specifically, $T_f$ determines the learned noise addition (see Eq.~\eqref{eq:ddim_fdp}); higher $T_f$ removes more source information (increasing flexibility for style transfer) but risks corrupting the anatomy, while $T_r$ determines the reconstruction steps.
Since Stage II target-specific fine-tuning depends on Stage I sampling quality, we perform a hyperparameter search using the frozen Stage I model to identify the trajectory that achieves the optimal global alignment.
We test varying steps ($T_f, T_r \in \{5,15, 25, 35, 45\}$) on the DWI-THP training set. 
To quantify ``global harmonization quality'' without a specific target, we calculate the mean inter-site WD and SSIM.
A lower WD indicates that the chosen trajectory ($T_f, T_r$) successfully bridges the distribution gaps between sites, providing a stable and consistent starting point for subsequent target adaptation.
As shown in Fig.~\ref{fig:para}, the combination of $T_f=35$ and $T_r=25$ achieves the lowest WD and highest SSIM. 
This confirms that this trajectory offers the optimal balance, erasing sufficient source style to allow harmonization while preserving essential anatomy. 




\begin{figure}[!t]
\setlength{\abovecaptionskip}{0pt}
\setlength{\belowcaptionskip}{0pt}
\setlength{\abovedisplayskip}{0pt}
\setlength{\belowdisplayskip}{0pt}
\centering
\includegraphics[width=1\columnwidth]{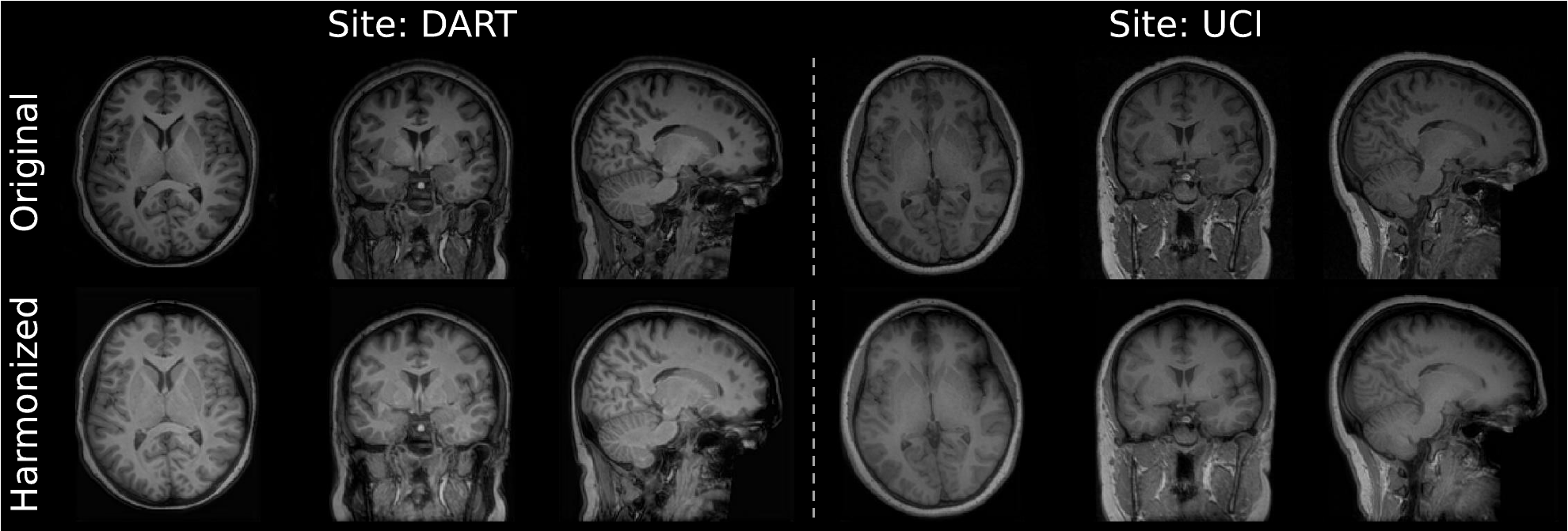}
\caption{
Zero-shot global harmonization on unprocessed MRIs. Visualization of raw inputs (Top) vs. globally harmonized outputs (Bottom) for Site DART and Site UCI across all three views (Axial, Coronal, Sagittal). 
}
\label{fig:unprocessed}
\end{figure}

\subsection{Influence of Preprocessing Steps}
Standard harmonization protocols~\cite{StyleGAN,ImUnity_2023,xu2024simix} typically treat image-level harmonization as the final stage of a preprocessing pipeline, operating on MRI volumes that have been skull-stripped, bias-corrected, and registered to standard templates (\eg,~MNI). 
To demonstrate robustness, we evaluate the zero-shot generalization by applying the frozen Stage I Global Harmonizer (trained on the preprocessed/skull-stripped data from our main experiment) directly to raw, unprocessed MRIs from DWI-THP test set. 
The visualization in Fig.~\ref{fig:unprocessed} shows the frozen model successfully projects the unprocessed MRIs from two sites into a unified intensity domain. 
Despite never observing skulls or background noise during training, it effectively harmonizes the brain tissue contrast while preserving the unseen non-brain structures, demonstrating MMH’s ability to operate on raw, native-space images without extensive preprocessing.

\subsection{Limitations and Future Work}
Several limitations of the current study need to be addressed. 
\emph{First}, this study validates harmonization and anatomical preservation on healthy cohorts. 
To establish broader clinical applicability, future work will extend validation to datasets with brain pathologies (\eg, Alzheimer's disease or Glioblastoma), ensuring that the structural fidelity observed in healthy brains generalizes effectively to anomalous anatomy and preserves disease-specific biomarkers. 
\emph{Second}, while scalable to multi-sequence data, the framework is bound to the discrete set of sequences seen during training. 
We plan to investigate zero-shot modality adaptation, potentially leveraging the semantic flexibility of the BiomedCLIP encoder to guide the model toward novel contrasts using text-based prompts or classifier-free guidance.
\emph{Additionally}, the reliance on voxel-space diffusion, while essential for avoiding the blurring artifacts introduced by latent compression, incurs higher computational costs than GANs or LDMs. 
Future optimizations will explore diffusion distillation~\cite{salimans2022progressive} to compress the generative process into a few sampling steps without compromising the high-frequency anatomical precision.

\section{Conclusion}
\label{S_conclusion}
We present MMH, a unified framework for multi-sequence and multi-site brain MRI harmonization. 
Integrating gradient-anchored diffusion with 3D semantic style priors (TPA-CLIP), MMH effectively decouples site-related style from biological content without paired training data. 
Extensive validation on over 4,000 MRI scans demonstrates significant improvements over state-of-the-art baselines in MRI harmonization. 
Crucially, MMH proves robust across multiple sequences (T1w/T2w) and processing pipelines, delivering superior style alignment while preserving the high-frequency anatomical details required for downstream tasks. 
By maintaining biological validity, from fine-grained subcortical segmentation to brain age-related features, MMH offers a trustworthy and scalable solution for large-scale, multi-institutional MRI analysis.

\if false 
\section*{Acknowledgment}
This research was supported in part by NIH grants (Nos. R01AG073297, RF1AG082938, R01EB035160, and R01NS134849). 
\fi


\footnotesize
\bibliography{reference}

@IEEEtranBSTCTL{IEEEexample:BSTcontrol,
  CTLuse_forced_etal       = "yes",
  CTLmax_names_forced_etal = "6",
  CTLnames_show_etal       = "1",
    CTLdash_repeated_names   = "no",
}

@inproceedings{isola2017image,
  title={Image-to-image translation with conditional adversarial networks},
  author={Isola, Phillip and Zhu, Jun-Yan and Zhou, Tinghui and Efros, Alexei A},
 booktitle={Proc. IEEE Conf. Comput. Vis. Pattern Recognit. (CVPR)},
  pages={1125--1134},
  year={2017}
}

@inproceedings{blau2018perception,
  title={The perception-distortion tradeoff},
  author={Blau, Yochai and Michaeli, Tomer},
 booktitle={Proc. IEEE Conf. Comput. Vis. Pattern Recognit. (CVPR)},
  pages={6228--6237},
  year={2018}
}

@article{odena2016deconvolution,
  title={Deconvolution and checkerboard artifacts},
  author={Odena, Augustus and Dumoulin, Vincent and Olah, Chris},
  journal={Distill},
  volume={1},
  number={10},
  pages={e3},
  year={2016}
}

@inproceedings{zhang2018translating,
  title={Translating and segmenting multimodal medical volumes with cycle-and shape-consistency generative adversarial network},
  author={Zhang, Zizhao and Yang, Lin and Zheng, Yefeng},
booktitle={Proc. IEEE Conf. Comput. Vis. Pattern Recognit. (CVPR)},
  pages={9242--9251},
  year={2018}
}

@article{vaswani2017attention,
  title={Attention is all you need},
  author={Vaswani, Ashish and Shazeer, Noam and Parmar, Niki and Uszkoreit, Jakob and Jones, Llion and Gomez, Aidan N and Kaiser, {\L}ukasz and Polosukhin, Illia},
  journal={Advances In Neural Information Processing Systems},
  volume={30},
  year={2017}
}

@article{nie2018medical,
  title={Medical image synthesis with deep convolutional adversarial networks},
  author={Nie, Dong and Trullo, Roger and Lian, Jun and Wang, Li and Petitjean, Caroline and Ruan, Su and Wang, Qian and Shen, Dinggang},
journal={IEEE Trans. Biomed. Eng.},
volume={65},
  number={12},
  pages={2720--2730},
  year={2018},
  publisher={IEEE}
}

@book{gonzalez2009digital,
  title={Digital image processing},
  author={Gonzalez, Rafael C},
  year={2009},
  publisher={Pearson Education India}
}

@article{salimans2022progressive,
  title={Progressive distillation for fast sampling of diffusion models},
  author={Salimans, Tim and Ho, Jonathan},
  journal={arXiv preprint arXiv:2202.00512},
  year={2022}
}

@article{fischl2012freesurfer,
  title={FreeSurfer},
  author={Fischl, Bruce},
  journal={NeuroImage},
  volume={62},
  number={2},
  pages={774--781},
  year={2012},
  publisher={Elsevier}
}

@article{fischl1999cortical,
  title={Cortical surface-based analysis {II}: inflation, flattening, and a surface-based coordinate system},
  author={Fischl, Bruce and Sereno, Martin I and Dale, Anders M},
  journal={NeuroImage},
  volume={9},
  number={2},
  pages={195--207},
  year={1999},
  publisher={Elsevier}
}

@article{ge2002age,
  title={{Age-related total gray matter and white matter changes in normal adult brain. Part I: volumetric MR imaging analysis}},
  author={Ge, Yulin and Grossman, Robert I and Babb, James S and Rabin, Marcie L and Mannon, Lois J and Kolson, Dennis L},
  journal={American Journal of Neuroradiology},
  volume={23},
  number={8},
  pages={1327--1333},
  year={2002},
  publisher={American Journal of Neuroradiology}
}

@article{maaten2008visualizing,
  title={{Visualizing data using t-SNE}},
  author={Maaten, Laurens van der and Hinton, Geoffrey},
  journal={Journal of Machine Learning Research},
  volume={9},
  number={Nov},
  pages={2579--2605},
  year={2008}
}

@article{kazeminia2020gans,
  title={{GANs for medical image analysis}},
  author={Kazeminia, Salome and Baur, Christoph and Kuijper, Arjan and Van Ginneken, Bram and Navab, Nassir and Albarqouni, Shadi and Mukhopadhyay, Anirban},
  journal={Artificial Intelligence in Medicine},
  volume={109},
  pages={101938},
  year={2020},
  publisher={Elsevier}
}

@inproceedings{cohen2018distribution,
  title={Distribution matching losses can hallucinate features in medical image translation},
  author={Cohen, Joseph Paul and Luck, Margaux and Honari, Sina},
  booktitle={Proc. Int. Conf. Med. Image Comput. Comput.-Assist. Interv. (MICCAI)},
  pages={529--536},
  year={2018},
  organization={Springer}
}

@article{wachinger2021detect,
  title={Detect and correct bias in multi-site neuroimaging datasets},
  author={Wachinger, Christian and Rieckmann, Anna and P{\"o}lsterl, Sebastian and Alzheimer’s Disease Neuroimaging Initiative and others},
  journal={Medical Image Analysis},
  volume={67},
  pages={101879},
  year={2021},
  publisher={Elsevier}
}

@incollection{durrer2023denoising,
  title={Denoising diffusion models for inpainting of healthy brain tissue},
  author={Durrer, Alicia and Cattin, Philippe C and Wolleb, Julia},
  booktitle={International Challenge on Cross-Modality Domain Adaptation for Medical Image Segmentation},
  pages={35--45},
  year={2023},
  publisher={Springer}
}

@article{vazquez2024review,
  title={A review of latent representation models in neuroimaging},
  author={V{\'a}zquez-Garc{\'\i}a, C and Mart{\'\i}nez-Murcia, FJ and Rom{\'a}n, F Segovia and G{\'o}rriz, Juan M},
  journal={arXiv preprint arXiv:2412.19844},
  year={2024}
}

@article{croitoru2023diffusion,
  title={Diffusion models in vision: A survey},
  author={Croitoru, Florinel-Alin and Hondru, Vlad and Ionescu, Radu Tudor and Shah, Mubarak},
    journal={IEEE Trans. Pattern Anal. Mach. Intell.},
  volume={45},
  number={9},
  pages={10850--10869},
  year={2023},
  publisher={Ieee}
}

@article{dewey2019deepharmony,
  title={DeepHarmony: A deep learning approach to contrast harmonization across scanner changes},
  author={Dewey, Blake E and Zhao, Can and Reinhold, Jacob C and Carass, Aaron and Fitzgerald, Kathryn C and Sotirchos, Elias S and Saidha, Shiv and Oh, Jiwon and Pham, Dzung L and Calabresi, Peter A and others},
  journal={Magnetic Resonance Imaging},
  volume={64},
  pages={160--170},
  year={2019},
  publisher={Elsevier}
}

@inproceedings{jung2021conditional,
  title={{Conditional GAN with an attention-based generator and a 3D discriminator for 3D medical image generation}},
  author={Jung, Euijin and Luna, Miguel and Park, Sang Hyun},
  booktitle={Proc. Int. Conf. Med. Image Comput. Comput.-Assist. Interv. (MICCAI)},
  pages={318--328},
  year={2021},
  organization={Springer}
}

@inproceedings{guan2022fast,
  title={Fast image-level {MRI} harmonization via spectrum analysis},
  author={Guan, Hao and Liu, Siyuan and Lin, Weili and Yap, Pew-Thian and Liu, Mingxia},
     booktitle={Proc. Int. Workshop Mach. Learn. Med. Imag. (MLMI)},
  pages={201--209},
  year={2022},
  organization={Springer}
}

@article{nyul2000new,
  title={New variants of a method of {MRI} scale standardization},
  author={Ny{\'u}l, L{\'a}szl{\'o} G and Udupa, Jayaram K and Zhang, Xuan},
  journal={IEEE Transactions on Medical Imaging},
  volume={19},
  number={2},
  pages={143--150},
  year={2000},
  publisher={IEEE}
}

@article{shinohara2014statistical,
  title={Statistical normalization techniques for magnetic resonance imaging},
  author={Shinohara, Russell T and Sweeney, Elizabeth M and Goldsmith, Jeff and Shiee, Navid and Mateen, Farrah J and Calabresi, Peter A and Jarso, Samson and Pham, Dzung L and Reich, Daniel S and Crainiceanu, Ciprian M and others},
  journal={NeuroImage: Clinical},
  volume={6},
  pages={9--19},
  year={2014},
  publisher={Elsevier}
}

@inproceedings{wu2025unpaired,
  title={Unpaired Multi-site Brain {MRI} Harmonization with Image Style-Guided Latent Diffusion},
  author={Wu, Mengqi and Yu, Minhui and Lin, Weili and Yap, Pew-Thian and Liu, Mingxia},
  booktitle={Proc. Int. Conf. Med. Image Comput. Comput.-Assist. Interv. (MICCAI)},
  pages={683--693},
  year={2025},
  organization={Springer}
}

@inproceedings{CLIP,
  title={Learning transferable visual models from natural language supervision},
  author={Radford, Alec and Kim, Jong Wook and Hallacy, Chris and Ramesh, Aditya and Goh, Gabriel and Agarwal, Sandhini and Sastry, Girish and Askell, Amanda and Mishkin, Pamela and Clark, Jack and others},
booktitle={Proc. Int. Conf. Mach. Learn. (ICML)},
pages={8748--8763},
  year={2021},
organization={PMLR}
}

@article{tian2022deep,
  title={A deep learning-based multisite neuroimage harmonization framework established with a traveling-subject dataset},
  author={Tian, Dezheng and Zeng, Zilong and Sun, Xiaoyi and Tong, Qiqi and Li, Huanjie and He, Hongjian and Gao, Jia-Hong and He, Yong and Xia, Mingrui},
  journal={NeuroImage},
  volume={257},
  pages={119297},
  year={2022},
  publisher={Elsevier}
}

@article{zuo2023haca3,
  title={{HACA3: A unified approach for multi-site MR image harmonization}},
  author={Zuo, Lianrui and Liu, Yihao and Xue, Yuan and Dewey, Blake E and Remedios, Samuel W and Hays, Savannah P and Bilgel, Murat and Mowry, Ellen M and Newsome, Scott D and Calabresi, Peter A and others},
  journal={Computerized Medical Imaging and Graphics},
  volume={109},
  pages={102285},
  year={2023},
  publisher={Elsevier}
}

@article{wu2025disentangled,
  title={{Disentangled latent energy-based style translation: An image-level structural MRI harmonization framework}},
  author={Wu, Mengqi and Zhang, Lintao and Yap, Pew-Thian and Zhu, Hongtu and Liu, Mingxia},
  journal={Neural Networks},
  volume={184},
  pages={107039},
  year={2025},
  publisher={Elsevier}
}

@inproceedings{zuo2022disentangling,
  title={{Disentangling a single MR modality}},
  author={Zuo, Lianrui and Liu, Yihao and Xue, Yuan and Han, Shuo and Bilgel, Murat and Resnick, Susan M and Prince, Jerry L and Carass, Aaron},
booktitle={Proc. Int. Workshop Data Augment. Label. Imperfect. (DALI)},
  pages={54--63},
  year={2022},
  organization={Springer}
}

@article{xu2024simix,
  title={{SiMix: A domain generalization method for cross-site brain MRI harmonization via site mixing}},
  author={Xu, Chundan and Li, Jie and Wang, Yakui and Wang, Lixue and Wang, Yizhe and Zhang, Xiaofeng and Liu, Weiqi and Chen, Jingang and Vatian, Aleksandra and Gusarova, Natalia and others},
  journal={NeuroImage},
  volume={299},
  pages={120812},
  year={2024},
  publisher={Elsevier}
}

@article{durrer2023diffusion,
  title={Diffusion models for contrast harmonization of magnetic resonance images},
  author={Durrer, Alicia and Wolleb, Julia and Bieder, Florentin and Sinnecker, Tim and Weigel, Matthias and Sandk{\"u}hler, Robin and Granziera, Cristina and Yaldizli, {\"O}zg{\"u}r and Cattin, Philippe C},
  journal={arXiv preprint arXiv:2303.08189},
  year={2023}
}

@inproceedings{beizaee2023HF,
  title={{Harmonizing Flows: Unsupervised MR harmonization based on normalizing flows}},
  author={Beizaee, Farzad and Desrosiers, Christian and Lodygensky, Gregory A and Dolz, Jose},
booktitle={Proc. Int. Conf. Inf. Process. Med. Imag. (IPMI)},
  pages={347--359},
  year={2023},
  organization={Springer}
}

@article{billot2023robust,
  title={{Robust machine learning segmentation for large-scale analysis of heterogeneous clinical brain MRI datasets}},
  author={Billot, Benjamin and Magdamo, Colin and Cheng, You and Arnold, Steven E and Das, Sudeshna and Iglesias, Juan Eugenio},
  journal={Proceedings of the National Academy of Sciences},
  volume={120},
  number={9},
  pages={e2216399120},
  year={2023},
  publisher={National Academy of Sciences}
}

@inproceedings{parida2024quantitative,
  title={Quantitative metrics for benchmarking medical image harmonization},
  author={Parida, Abhijeet and Jiang, Zhifan and Packer, Roger J and Avery, Robert A and Anwar, Syed M and Linguraru, Marius G},
  booktitle={Proc. IEEE Int. Symp. Biomed. Imaging (ISBI)},
  pages={1--5},
  year={2024},
}

@article{song2020ddim,
  title={Denoising diffusion implicit models},
  author={Song, Jiaming and Meng, Chenlin and Ermon, Stefano},
  journal={arXiv preprint arXiv:2010.02502},
  year={2020}
}

@article{zhang2024biomedclip,
  title={A Multimodal Biomedical Foundation Model Trained from Fifteen Million Image–Text Pairs},
  author={Sheng Zhang and Yanbo Xu and Naoto Usuyama and Hanwen Xu and Jaspreet Bagga and Robert Tinn and Sam Preston and Rajesh Rao and Mu Wei and Naveen Valluri and Cliff Wong and Andrea Tupini and Yu Wang and Matt Mazzola and Swadheen Shukla and Lars Liden and Jianfeng Gao and Angela Crabtree and Brian Piening and Carlo Bifulco and Matthew P. Lungren and Tristan Naumann and Sheng Wang and Hoifung Poon},
  journal={NEJM AI},
  year={2024},
  volume={2},
  number={1}
}

@article{cardoso2022monai,
  title={Monai: An open-source framework for deep learning in healthcare},
  author={Cardoso, M Jorge and Li, Wenqi and Brown, Richard and Ma, Nic and Kerfoot, Eric and Wang, Yiheng and Murrey, Benjamin and Myronenko, Andriy and Zhao, Can and Yang, Dong and others},
  journal={arXiv preprint arXiv:2211.02701},
  year={2022}
}

@article{magnotta2012multicenter,
  title={Multicenter reliability of diffusion tensor imaging},
  author={Magnotta, Vincent A and Matsui, Joy T and Liu, Dawei and Johnson, Hans J and Long, Jeffrey D and Bolster Jr, Bradley D and Mueller, Bryon A and Lim, Kelvin and Mori, Susumu and Helmer, Karl G and others},
  journal={Brain Connectivity},
  volume={2},
  number={6},
  pages={345--355},
  year={2012},
  publisher={Mary Ann Liebert, Inc. 140 Huguenot Street, 3rd Floor New Rochelle, NY 10801 USA}
}

@article{glocker2019machine,
  title={Machine learning with multi-site imaging data: An empirical study on the impact of scanner effects},
  author={Glocker, Ben and Robinson, Robert and Castro, Daniel C and Dou, Qi and Konukoglu, Ender},
  journal={arXiv preprint arXiv:1910.04597},
  year={2019}
}

@article{gadewar2024synthesizing,
  title={Synthesizing study-specific controls using generative models on open access datasets for harmonized multi-study analyses},
  author={Gadewar, Shruti P and Zhu, Alyssa H and Gari, Iyad Ba and Somu, Sunanda and Thomopoulos, Sophia I and Thompson, Paul M and Nir, Talia M and Jahanshad, Neda},
  journal={arXiv preprint arXiv:2403.00093},
  year={2024}
}

@article{schnack2010mapping,
  title={{Mapping reliability in multicenter MRI: Voxel-based morphometry and cortical thickness}},
  author={Schnack, Hugo G and van Haren, Neeltje EM and Brouwer, Rachel M and van Baal, G Caroline M and Picchioni, Marco and Weisbrod, Matthias and Sauer, Heinrich and Cannon, Tyrone D and Huttunen, Matti and Lepage, Claude and others},
  journal={Human Brain Mapping},
  volume={31},
  number={12},
  pages={1967--1982},
  year={2010},
  publisher={Wiley Online Library}
}

@article{an2022goal,
  title={{Goal-specific brain MRI harmonization}},
  author={An, Lijun and Chen, Jianzhong and Chen, Pansheng and Zhang, Chen and He, Tong and Chen, Christopher and Zhou, Juan Helen and Yeo, BT Thomas},
  journal={NeuroImage},
  volume={263},
  pages={119570},
  year={2022},
  publisher={Elsevier}
}

@article{tofts2011multicentre,
  title={Multicentre imaging measurements for oncology and in the brain},
  author={Tofts, PS and Collins, DJ},
  journal={The British Journal of Radiology},
  volume={84},
  pages={S213--S226},
  year={2011}
}

@article{CALAMITI_2021,
  title={{Unsupervised MR harmonization by learning disentangled representations using information bottleneck theory}},
  author={Zuo, Lianrui and Dewey, Blake E and Liu, Yihao and He, Yufan and Newsome, Scott D and Mowry, Ellen M and Resnick, Susan M and Prince, Jerry L and Carass, Aaron},
  journal={NeuroImage},
  volume={243},
  pages={118569},
  year={2021},
  publisher={Elsevier}
}

@article{ImUnity_2023,
  title={{ImUnity: A generalizable VAE-GAN solution for multicenter MR image harmonization}},
  author={Cackowski, Stenzel and Barbier, Emmanuel L and Dojat, Michel and Christen, Thomas},
  journal={Medical Image Analysis},
  volume={88},
  pages={102799},
  year={2023},
  publisher={Elsevier}
}

@inproceedings{dewey2020disentangled,
  title={{A disentangled latent space for cross-site MRI harmonization}},
  author={Dewey, Blake E and Zuo, Lianrui and Carass, Aaron and He, Yufan and Liu, Yihao and Mowry, Ellen M and Newsome, Scott and Oh, Jiwon and Calabresi, Peter A and Prince, Jerry L},
  booktitle={{Proc. Int. Conf. Med. Image Comput. Comput.-Assist. Interv. (MICCAI)}},
  pages={720--729},
  year={2020},
  organization={Springer}
}

@inproceedings{Modanwal_2020_cyclegan,
  title={{MRI image harmonization using cycle-consistent generative adversarial network}},
  author={Modanwal, Gourav and Vellal, Adithya and Buda, Mateusz and Mazurowski, Maciej A},
  booktitle={Computer-Aided Diagnosis},
  volume={11314},
  pages={259--264},
  year={2020},
  organization={SPIE}
}

@article{chang_2022_cyclegan,
  title={Self-supervised learning for multi-center magnetic resonance imaging harmonization without traveling phantoms},
  author={Chang, Xiao and Cai, Xin and Dan, Yibo and Song, Yang and Lu, Qing and Yang, Guang and Nie, Shengdong},
  journal={Physics in Medicine \& Biology},
  volume={67},
  number={14},
  pages={145004},
  year={2022},
  publisher={IOP Publishing}
}

@inproceedings{StyleGAN,
  title={{Style transfer using generative adversarial networks for multi-site MRI harmonization}},
  author={Liu, Mengting and Maiti, Piyush and Thomopoulos, Sophia and Zhu, Alyssa and Chai, Yaqiong and Kim, Hosung and Jahanshad, Neda},
  booktitle={Proc. Int. Conf. Med. Image Comput. Comput.-Assist. Interv. (MICCAI)},
  pages={313--322},
  year={2021},
  organization={Springer}
}

@inproceedings{gatys2016style,
  title={Image style transfer using convolutional neural networks},
  author={Gatys, Leon A and Ecker, Alexander S and Bethge, Matthias},
 booktitle={Proc. IEEE Conf. Comput. Vis. Pattern Recognit. (CVPR)},
  pages={2414--2423},
  year={2016}
}

@inproceedings{huang2017arbitrary,
  title={Arbitrary style transfer in real-time with adaptive instance normalization},
  author={Huang, Xun and Belongie, Serge},
  booktitle={Proc. IEEE Int. Conf. Comput. Vis. (ICCV)},
  pages={1501--1510},
  year={2017}
}

@article{ho2020DDPM,
  title={Denoising diffusion probabilistic models},
  author={Ho, Jonathan and Jain, Ajay and Abbeel, Pieter},
  journal={Advances in Neural Information Processing Systems},
  volume={33},
  pages={6840--6851},
  year={2020}
}

@article{OpenBHB,
      title={{OpenBHB: A large-scale multi-site brain MRI data-set for age prediction and debiasing}},
      author={Dufumier, Benoit and Grigis, Antoine and Victor, Julie and Ambroise, Corentin and Frouin, Vincent and Duchesnay, Edouard},
        volume={263},
  pages={119637},
      journal={NeuroImage},
      year={2022}
}

@article{SRPBS_TS,
author = {Tanaka, Saori and Yamashita, Ayumu and Yahata, Noriaki and Itahashi, Takashi and Lisi, Giuseppe and Yamada, Takashi and Ichikawa, Naho and Takamura, Masahiro and Yoshihara, Yujiro and Kunimatsu, Akira and Okada, Naohiro and Hashimoto, Ryuichiro and Okada, Go and Sakai, Yuki and Morimoto, Jun and Narumoto, Jin and Shimada, Yasuhiro and Mano, Hiroaki and Yoshida, Wako and Imamizu, Hiroshi},
year = {2021},
pages = {227},
title = {A multi-site, multi-disorder resting-state magnetic resonance image database},
  volume={8},
  number={1},
  pages={227},
journal = {Scientific Data},
}

@inproceedings{wang2023stylediffusion,
  title={Stylediffusion: Controllable disentangled style transfer via diffusion models},
  author={Wang, Zhizhong and Zhao, Lei and Xing, Wei},
booktitle={Proc. IEEE Int. Conf. Comput. Vis. (ICCV)},
  pages={7677--7689},
  year={2023}
}

@misc{pomponio2020harmonization,
  title={{Harmonization of large MRI datasets for the analysis of brain imaging patterns throughout the lifespan}},
  author={Pomponio, R and Erus, G and Habes, M and Doshi, J and Srinivasan, D and Mamourian, E and Bashyam, V and Nasrallah, IM and Satterthwaite, TD and Fan, Y and others},  
  journal={NeuroImage},
  volume = {208},
  pages ={116450}, 
  year={2020}
}

@article{neuroCombat,
  title={Harmonization of cortical thickness measurements across scanners and sites},
  author={Fortin, Jean-Philippe and Cullen, Nicholas and Sheline, Yvette I and Taylor, Warren D and Aselcioglu, Irem and Cook, Philip A and Adams, Phil and Cooper, Crystal and Fava, Maurizio and McGrath, Patrick J and others},
  journal={NeuroImage},
  volume={167},
  pages={104--120},
  year={2018},
  publisher={Elsevier}
}

@article{ComBat_GAM,
  title={H{armonization of large MRI datasets for the analysis of brain imaging patterns throughout the lifespan}},
  author={Pomponio, Raymond and Erus, Guray and Habes, Mohamad and Doshi, Jimit and Srinivasan, Dhivya and Mamourian, Elizabeth and Bashyam, Vishnu and Nasrallah, Ilya M and Satterthwaite, Theodore D and Fan, Yong and others},
  journal={NeuroImage},
  volume={208},
  pages={116450},
  year={2020},
}

@inproceedings{CycleGAN2017,
  title={Unpaired image-to-image translation using cycle-consistent adversarial networks},
  author={Zhu, Jun-Yan and Park, Taesung and Isola, Phillip and Efros, Alexei A},
  booktitle={Proc. IEEE Int. Conf. Comput. Vis. (ICCV)},
  pages={2223--2232},
  year={2017}
}
\bibliographystyle{IEEEtran}

\end{document}